\documentclass{article} 
\usepackage{iclr2025_conference,times}

\usepackage{amsmath,amsfonts,bm}

\def\eqref#1{equation~\ref{#1}}

\def\1{\bm{1}}

\DeclareMathAlphabet{\mathsfit}{\encodingdefault}{\sfdefault}{m}{sl}
\SetMathAlphabet{\mathsfit}{bold}{\encodingdefault}{\sfdefault}{bx}{n}

\usepackage{hyperref}
\usepackage{url}
\usepackage{inconsolata}
\usepackage{subfigure}
\usepackage{wrapfig}
\usepackage{pmboxdraw}

\usepackage{multirow}
\usepackage{graphicx}
\usepackage{booktabs}
\usepackage{tabularx}
\usepackage{CJKutf8}
\usepackage{amssymb}
\usepackage{amsmath}
\usepackage{tabu} 
\usepackage{hyperref}
\usepackage{breakurl}
\usepackage{times}
\usepackage{latexsym}
\usepackage{tikz}
\usepackage{pgfplots}
\usepackage[normalem]{ulem}
\usepackage{xcolor}
\usepackage{dashrule}
\usepackage{colortbl}
\usepackage{tipa}
\usepackage{caption}
\usepackage{subcaption}
\usepackage{amssymb}
\usepackage{fontawesome}
\usepackage{tipa}
\usepackage{xspace}
\usepackage{pifont}%
\usepackage[ruled, lined, linesnumbered, longend]{algorithm2e}
\usepackage{ulem}
\usepackage{arydshln}

\definecolor{brickred}{HTML}{b92622}
\definecolor{midnightblue}{HTML}{005c7f}
\definecolor{salmon}{HTML}{f1958d}
\definecolor{burntorange}{HTML}{f19249}
\definecolor{junglegreen}{HTML}{4dae9d}
\definecolor{forestgreen}{HTML}{499c5e}
\definecolor{pinegreen}{HTML}{3d8a75}
\definecolor{seagreen}{HTML}{1EC652}
\definecolor{limegreen}{HTML}{97c65a}
\definecolor{redtable}{HTML}{E67F72}
\definecolor{orangetable}{HTML}{E67F72}
\definecolor{greentable}{HTML}{E67F72}
\definecolor{mypink}{HTML}{f4919d}
\definecolor{myred}{HTML}{e74f51}
\definecolor{mygreen}{HTML}{00b050}

\usepackage{lipsum}

\newcommand{\rnum}[1]{\uppercase\expandafter{\romannumeral #1\relax}}

\newcommand{\model}[1]{\textsc{#1}\xspace}
\newcommand{\sft}{\model{SFT}}
\newcommand{\iti}{\model{ITI}}
\newcommand{\caa}{\model{CAA}}
\newcommand{\llama}{\model{LLaMA2-7b-chat}}
\newcommand{\bloomz}{\model{BLOOMZ-7b}}
\newcommand{\mistral}{\model{Mistral-7b}}
\newcommand{\falcon}{\model{Falcon-7b-instruct}}
\newcommand{\bloomzfive}{\model{BLOOMZ-560m}}
\newcommand{\bloomzone}{\model{BLOOMZ-1b}}
\newcommand{\bloomzthree}{\model{BLOOMZ-3b}}
\newcommand{\baseline}{\model{baseline}}
\newcommand{\sadi}{\model{SADI}}
\newcommand{\sadihead}{\model{SADI-Head}}
\newcommand{\sadihidden}{\model{SADI-Hidden}}
\newcommand{\sadineuron}{\model{SADI-Neuron}}

\newcommand{\dataset}[1]{\texttt{#1}\xspace}
\newcommand{\mmlu}{\dataset{MMLU}}
\newcommand{\copa}{\dataset{COPA}}
\newcommand{\story}{\dataset{StoryCloze}}
\newcommand{\nli}{\dataset{NLI}}
\newcommand{\ssttwo}{\dataset{SST2}}
\newcommand{\sstfive}{\dataset{SST5}}
\newcommand{\boolq}{\dataset{BoolQ}}
\newcommand{\wino}{\dataset{Winogrande}}
\newcommand{\trivia}{\dataset{TriviaQA}}
\newcommand{\toxi}{\dataset{ToxiGen}}
\newcommand{\truthful}{\dataset{TruthfulQA}}
\newcommand{\xcopa}{\dataset{XCOPA}}

\usepackage[T1]{fontenc}

\usepackage[utf8]{inputenc}

\usepackage{microtype}

\title{Semantics-Adaptive Activation Intervention for LLMs via Dynamic Steering Vectors}

\author{%
Weixuan Wang\textsuperscript{1}  \quad Jingyuan Yang\textsuperscript{2}  \quad
Wei Peng\textsuperscript{3}
 \\[1ex]
\textsuperscript{1}School of Informatics, University of Edinburgh  \quad
\textsuperscript{2}Huawei Technologies Co., Ltd. \\
\textsuperscript{3}School of Engineering, RMIT University \\
\texttt{weixuan.wang@ed.ac.uk}   \quad
\texttt{yangjingyuan2@huawei.com}   \quad
\texttt{wei.peng3@rmit.edu.au}
}

\iclrfinalcopy 
\begin{document}

\maketitle

\begin{abstract}
Large language models (LLMs) have achieved remarkable performance across many tasks, yet aligning them with desired behaviors remains challenging. Activation intervention has emerged as an effective and economical method to modify the behavior of LLMs. Despite considerable interest in this area, current intervention methods exclusively employ a fixed steering vector to modify model activations, lacking adaptability to diverse input semantics. To address this limitation, we propose \textbf{Semantics-Adaptive Dynamic Intervention (\sadi)}, a novel method that constructs a dynamic steering vector to intervene model activations at inference time. More specifically, \sadi utilizes activation differences in contrastive pairs to precisely identify critical elements of an LLM (i.e., attention heads, hidden states, and neurons) for targeted intervention. During inference, \sadi dynamically steers model behavior by scaling element-wise activations based on the directions of input semantics. Experimental results show that \sadi outperforms established baselines by substantial margins, improving task performance without training. \sadi's cost-effectiveness and generalizability across various LLM backbones and tasks highlight its potential as a versatile alignment technique.\footnote{https://github.com/weixuan-wang123/SADI}

\end{abstract}

\section{Introduction}

Large language models (LLMs) have demonstrated remarkable capabilities across many tasks \citep{gpt4,llama,palm,gemini,gemma}. Nevertheless, aligning these models to target behaviors remains challenging \citep{flan,instruction}. Existing approaches like supervised fine-tuning \citep{sft} (SFT), Reinforcement Learning from Human Feedback \citep{rhlf} (RLHF), and prompt engineering \citep{prompt-engineering,wang2023retrieval} are effective but have limitations. They often require extensive datasets, struggle to prevent hallucinations, and sometimes fail to produce the desired results. 

Recently, advancements in model alignment techniques, known as ``activation engineering'', aim to address these limitations \citep{engineering1,engineering2,engineering3,iti,trfr}. Activation engineering involves making targeted modifications to the internal activations of LLMs to guide their outputs more precisely. This technique constructs steering vectors that, when integrated into the forward pass of a frozen LLM, induce specific desirable changes in the output text. However, traditional steering vectors are static and may not adapt well to the diverse semantic contexts encountered during inference \citep{addition,steer-vector}. This misalignment between the direction of steering vector and the input's semantic direction can adversely impact the model's predictive performance, particularly when the discrepancy is substantial. These limitations highlight the need for dynamic and adaptive steering mechanisms capable of effectively handling varied input semantics. 

In this work, we introduce the \textbf{Semantics-Adaptive Dynamic Intervention} (\textbf{\sadi}), a novel approach designed to overcome the limitations of fixed steering mechanisms. \sadi adjusts model activations by dynamically generating a steering vector tailored to each input's semantic context. Specifically, \sadi utilizes activation differences from contrastive pairs to create a binary mask that identifies critical model elements for targeted intervention. During inference, this mask is applied to user input activations with element-wise scaling, effectively manipulating the LLM's behavior to align with the input semantics. This process ensures that modifications preserve the semantic alignment of the inputs, allowing for more precise and context-sensitive interventions. Furthermore, we apply \sadi to various components of LLMs, including hidden states (\sadihidden), attention heads (\sadihead), and neurons in feed-forward networks (FFNs) (\sadineuron). 

To validate the effectiveness of \sadi, we conduct extensive experiments using four diverse model backbones: \llama, \bloomz, \mistral, \falcon across eleven widely used benchmarks. The experiments involve a comprehensive range of tasks, from multiple-choice tasks (\copa, \story, \nli, \mmlu, \ssttwo, \sstfive, \boolq, \wino), to open-ended generation tasks (\trivia, \toxi, and \truthful). Our experimental results reveal that \sadi significantly outperforms existing activation intervention methods.

Our contributions are summarized as follows:
\begin{itemize}
    \item We propose a dynamic activation intervention approach named Semantics-Adaptive Dynamic Intervention (\sadi), which automatically modulates LLM activations at inference time to adapt to varied input semantics without requiring any additional training (see \autoref{sec:method}).

    \item \sadi is a generic steering method applicable to a wide range of LLMs. Through extensive experiments with four model backbones over eleven diverse tasks, \sadi has proven to significantly enhance model performance, surpassing baseline methods by substantial margins, with accuracy improvements reaching up to +14.69 (see \autoref{sec:experiments}). Our detailed analysis demonstrates that interventions targeting attention heads (\sadihead) consistently yields significant performance improvements across various tasks, validating the effectiveness of our dynamic steering approach (see \autoref{sec:experiments}). 

    \item  \sadi demonstrates excellent generalizability across different model sizes, few-shot settings, and multilingual scenarios (see \autoref{sec:discussion}).
    We further show that \sadi is a cost-effective steering method that necessitates only a small number of additional examples (i.e., 150 items) in developing a dynamic steering vector and does not require any training (see \autoref{sec:analysis}).
\end{itemize}

\begin{figure*}[t]
    \centering
    \includegraphics[scale=0.35]{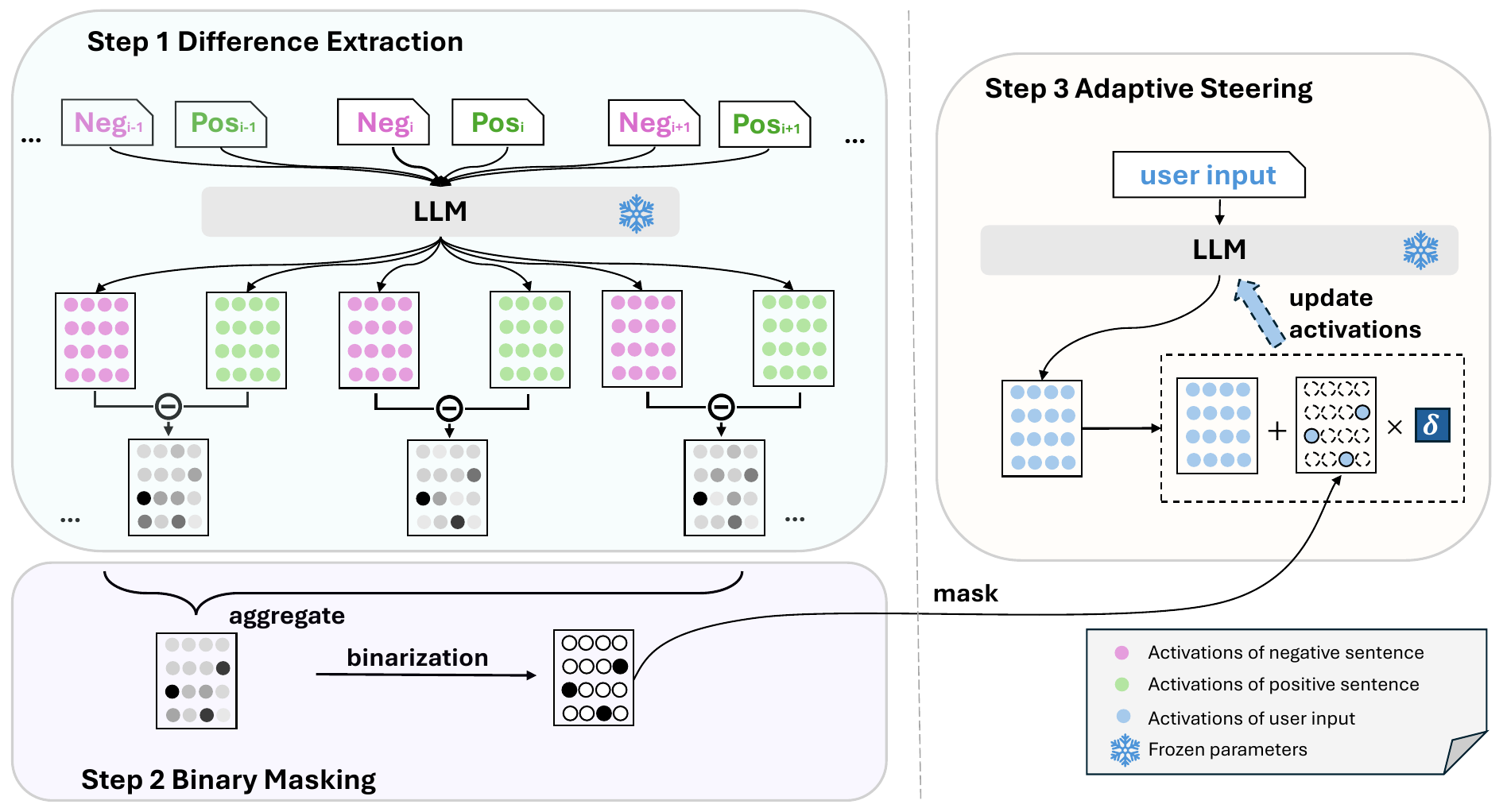}
    \caption{Three steps of \sadi: (1) Difference Extraction: extract the activation differences between positive and negative examples from all model layers; (2) Binary Masking: compute the mean activation difference to locate the key elements and produce an identification mask by binarization; and (3) Adaptive Steering: intervene the activations during inference by applying the identification mask to the input activations scaled by a factor of $\delta$.}
    \label{framework}
\end{figure*}

\section{Related Work}
Activation engineering has been proposed as a cost-effective method to modify an LLM's activations during decoding \citep{engineering2, engineering3,wang2024bridging}. By analyzing activation differences between contrastive pairs, these methods identify specific directions to adjust LLM behaviors. For instance, some studies have improved LLM truthfulness by shifting activations along vectors between true and false output distributions \citep{iti, trfr}. Moreover, discrepancies in contextual examples have been used to identify crucial modifications needed to reduce LLM toxicity \citep{context-vector}.
More generally, these differences can be used to update the residual stream without requiring explicit direction settings for adjustments. \citet{addition} construct steering vectors by assessing intermediate activation differences between two prompts, effectively shifting emotions from negative to positive. 
Similarly, \citet{steer-vector} use contrast pairs to create steering vectors that modify model behaviors by adjusting hidden states.

Our problem formulation aligns with the ``linear representation hypothesis'' \citep{linear, addition} which posits that
high-level concepts (i.e., features of the input) are represented linearly as directions in the representation space. Model behavior intervention can be achieved by adding an appropriate steering vector to the representation of a concept without altering other concepts. One key insight is that intervening along the direction of the feature representation of an input is expected to enhance the probability of producing a desirable output. Intuitively, we need to preserve the semantic direction of an input when applying a steering vector to update activations. 

Existing methods use fixed steering vectors generated from additional contrastive pairs during intervention without aligning with input semantics \citep{addition,steer-vector}.  As activation patterns can vary significantly across different inputs for the same task \citep{wang2024assessing,sharing}, model steering with a fixed vector may cause the intervention direction to deviate from the representation of input contexts. This highlights the need for more adaptive approaches. Developing a dynamic steering mechanism to adapt to the input semantics is critical for effective intervention. 

\section{Semantics-Adaptive Dynamic Intervention}
\label{sec:method}

In this section, we provide a comprehensive description of the proposed \sadi. We begin with an overview of \sadi in \autoref{sec:overview}, and then introduce each step of \sadi, including Difference Extraction (\autoref{sec:collect}), Binary Masking (\autoref{sec:identify}), and Adaptive Steering (\autoref{sec:intervene}).

\subsection{Overview of \sadi}
\label{sec:overview}
Our method, \sadi, encompasses three pivotal steps to dynamically steer model behavior, as shown in \autoref{framework} and \autoref{alg:sadi}. First, the activation differences between positive and negative examples are extracted across all layers of the model. These differences are aggregated to compute the mean difference, which is used to identify critical elements influencing the model’s behavior. Based on this computation, we create an identification mask through binarization, keeping the crucial elements while masking out the insignificant ones. Furthermore,  this mask is applied to the activations of user inputs, scaled by a factor during inference. In this way, we manage to manipulate the behaviors of LLMs. We present more details in the following sections.

\subsection{Difference Extraction}
\label{sec:collect}
In the initial step of \sadi, our objective is to extract activation differences from contrastive pairs to isolate the internal activations most closely associated with the target behavior of the language model. Specifically, we aim to identify features that distinguish positive outcomes (e.g., correct answers) from negative ones (e.g., incorrect answers), thereby allowing adjustments to the model's behavior.

Let $ \mathcal{P} = \{\mathcal{P}^{l} \mid 0 \leq l < L\} $ represents an LLM consisting of $ L $ layers, whose behavior we seek to modify. We build a dataset $ T = \{(x_i, y^{\textrm{pos}}_i, y^{\textrm{neg}}_i)\}_{i=1}^{N} $ containing $ N $ instances, where each instance includes the question $ x_i $, a positive output $ y^{\textrm{pos}}_i $, and a negative output $ y^{\textrm{neg}}_i $. For each instance $ i $ and layer $ (l) $, we obtain activations by forwarding the concatenation of the input and the corresponding output through the model $ \mathcal{P} $. Specifically, we forward $ x_i $ concatenated with $ y_i^{\text{pos}} $ to derive the positive activation $ \bm{\mathcal{A}}_{i, j}^{\text{pos}, (l)} = \mathcal{P}^{l} \left( x_i \; || \; y_i^{\text{pos}} \right)_j $, where $ \cdot || \cdot $ denotes concatenation and $ j $ represents the $ j $-th token in the sequence. Similarly, we obtain $ \bm{\mathcal{A}}_{i, j}^{\text{neg}, (l)} $ for the negative activation. Additionally, we focus on the activation of the last token in each sequence, as it typically encapsulates the complete semantics of the input-output pair. For simplicity, we denote the activation of the last token as $ \bm{\mathcal{A}}_{i}^{\text{pos}, (l)} $ and $ \bm{\mathcal{A}}_{i}^{\text{neg}, (l)} $ for the positive and negative outputs, respectively.

To identify the features within the model that differentiate correct from incorrect outputs, we compute the difference between the positive and negative activations for each instance at each layer as follows:
\begin{align}
    D_i^{(l)} = \bm{\mathcal{A}}^{\textrm{pos},(l)}_{i} - \bm{\mathcal{A}}^{\textrm{neg},(l)}_{i}.
\end{align}
By examining differences $ D_i^{(l)} $, we can determine which activations are crucial for model's behavior.

\subsection{Binary Masking}
\label{sec:identify}

We now present how we construct a mask to identify and focus interventions on the critical model elements that affect model's behavior. As shown in \autoref{framework} (Step 2), after extracting the activation differences from contrastive pairs, we compute the mean difference across all instances and layers, and concatenate them to build the overall mean difference vector $ D $ for all model elements:
\begin{align}
    D =  \textrm{Concat}(D^{(0)},D^{(1)},...,D^{(L-1)}) \textrm{, where } D^{(l)} = \frac{1}{N} \sum_{i=1}^{N} D_i^{(l)}.
    \label{eq:difference}
\end{align}
Here, $ D \in \mathbb{R}^{L \times d_m} $ represents the concatenated mean activation differences across all $ L $ layers, and $ d_m $ denotes the dimensionality of the model components, which may correspond to hidden states, attention heads, or neurons in FFNs.

We then binarize the mean activation difference $ D $ to create an identification mask $ M \in \mathbb{R}^{L \times d_m} $. This is done by setting the entries corresponding to the top-K elements with the largest differences to 1 and the rest to 0:
\begin{align}
    M[l, m] = \begin{cases}1 & (l, m) \in E_K \\
    0 & \text{otherwise}\end{cases},
\end{align}
where $ l $ indexes the layers, $ m $ indexes the model elements within a layer, and $ E_K $ is the set of indices of the top-K elements with the highest mean activation differences. This step ensures that \sadi focuses on the most impactful elements contributing to the desired behavior, reducing unnecessary alterations to non-essential elements and enhancing the efficiency of the intervention.

\subsection{Adaptive Steering}
\label{sec:intervene}
Previous activation intervention methods modify all activation elements indiscriminately, which can disrupt the model's overall behavior \citep{iti, steer-vector}. To mitigate this issue, we perform a focused intervention on the top-K elements during inference. By leaving irrelevant activations intact, our intervention becomes less intrusive and preserves the model's non-target behaviors. With this design, \sadi precisely adjusts activations to minimize disruption to the model's residual functionalities. This approach is visualized in Step 3 of \autoref{framework}.

Unlike previous studies using the fixed steering vector to intervene models, and inspired by \citet{linear}, we design a steering mechanism that considers the semantic direction of the input. The steering vector dynamically adapts to the input's semantic direction (see \autoref{eq:intervene}), maintaining the effectiveness of the intervention without deviating from the intended semantics.

For a given user input $q$, we first extract the activations of the last token from each layer: $\bm{\mathcal{A}}_{q}^{(l)}$ for $l = 0, 1, ..., L-1$. We then concatenate these activations to form a single vector:
\begin{align}
    \bm{\mathcal{A}}_{q} = \textrm{Concat}(\bm{\mathcal{A}}_{q}^{(0)}, \bm{\mathcal{A}}_{q}^{(1)}, ..., \bm{\mathcal{A}}_{q}^{(L-1)}).
\end{align}
Next, we apply the identification mask $M$ to these activations and update them using:
\begin{align}
    \bm{\mathcal{A}'}_{q} =\bm{\mathcal{A}}_{q} + \delta (\bm{\mathcal{A}}_{q} \odot M)
    \label{eq:intervene} .
\end{align}
Here, $\odot$ represents the element-wise product, and $\delta$ is a hyperparameter controlling the strength of the intervention along the input's semantic direction. By calculating the steering vector based on the activations of input $q$, the intervention dynamically aligns with the input's semantics. This approach maintains the direction of the activation projections, ensuring that the intervention remains semantically relevant and effective. Subsequently, we complete the altered forward pass with the updated activations $\bm{\mathcal{A}'}_{q}$.

\paragraph{Hyperparameters $K$ and $\delta$}Our method introduces two key hyperparameters: $K \in \mathbb{N}^+$, specifying the number of top elements targeted during the intervention, and $\delta \in \mathbb{R}^+$, controlling the strength of the intervention. We perform a hyperparameter sweep to empirically determine their optimal values. Detailed analysis of the hyperparameter selection is provided in \autoref{sec:main-results}.

By selectively targeting the most impactful activation elements and adapting our intervention to the input's semantic direction, our method effectively steers the model toward the desired behavior.

{
\setlength{\algomargin}{12pt}
\begin{algorithm}[H] \small
    \SetKwInOut{Input}{Input}
    \SetKwInOut{Output}{Output}
    \Input{$T = \{ (x^{\textrm{pos}}_i, x^{\textrm{neg}}_i) \}_{i=1}^{N}$, a set of contrastive pairs; $ U = \{ u_j \}_{j=1}^{K}$, a test set; $ \mathcal{P}$, a pre-trained LLM; $ A$, a function to extract activations from $ \mathcal{P}$;}
    \Output{$ O $, the modified outputs collection;}
    \SetAlgoLined

    $\Delta \mathcal{A} \leftarrow \mathbf{0}$  \tcp*[f]{Initialize the mean difference} \\
    \For(\tcp*[f]{Collect and compute the mean difference of activations}){i=1 to N}{
         $\Delta \mathcal{A} \leftarrow \Delta \mathcal{A} + (A(\mathcal{P}(x^{\textrm{pos}}_i)) - A(\mathcal{P}(x^{\textrm{neg}}_i)))$
    }
    $\Delta \mathcal{A} \leftarrow \frac{1}{N} \Delta \mathcal{A}$ \\
    $M \leftarrow \textrm{binarize}(\Delta \mathcal{A})$  \tcp*[f]{Create the identification mask} \\
    $O \leftarrow []$ \tcp*[f]{Intervene generation} \\
    \For{j=1 to K}{
        $\mathcal{A}_j \leftarrow A(P(u_j))$  \tcp*[f]{Extract activations for each input} \\
        $S_j \leftarrow M \odot \mathcal{A}_j$  \tcp*[f]{Apply mask to activations and update} \\
        $\mathcal{A}_j' \leftarrow \mathcal{A}_j + \delta \times S_j$ \\
        $O \leftarrow O \bigcup \{ P(\mathcal{A}_j') \}$ \tcp*[f]{Complete the modified forward pass}
    }
    \Return{$O$}
    \caption{\sadi: Semantics-Adaptive Dynamic Intervention}
    \label{alg:sadi}
\end{algorithm}
}

\section{Experiments}
\label{sec:experiments}
We present our experimental setup (\autoref{sec:experiment-settings})
, comparative methods (\autoref{sec:experimental-comparisons}), main results (\autoref{sec:main-results}), and specific contributions of \sadi's constitutes (\autoref{sec:scrutinize}) in this section.

\subsection{Experiment Settings}
\label{sec:experiment-settings}

In this subsection, we describe the experimental settings for evaluating \sadi. First, we outline the tasks and the evaluation metrics used to assess model performance. Following that, we detail the construction of contrastive pairs. Finally, we introduce the selected LLMs.

\paragraph{Tasks and Evaluation Metrics} We conduct experiments on the following two types of tasks: multiple-choice tasks and open-ended generation tasks. For the multiple-choice tasks, we use datasets: \copa~\citep{copa}, \story~\citep{storycloze}, \nli~\citep{nli}, \mmlu~\citep{mmlu}, \ssttwo~\citep{sst}, \sstfive~\citep{sst}, \boolq~\citep{boolq}, and \wino~\citep{winogrande}, with response formats ranging from 2-way to 5-way choices. Detailed descriptions of the datasets are provided in Appendix~\ref{appendix-data}. We measure performance across these tasks using accuracy.

For the open-ended generation tasks, we apply \sadi on \trivia~\citep{triviaqa}, \truthful~\citep{truthfulqa}, \toxi~\citep{toxigen} datasets. The Exact Match (EM) metric assesses \trivia, while \truthful is evaluated using multiple-choice accuracy (MC) and pre-trained judge models for truthfulness\footnote{https://huggingface.co/allenai/truthfulqa-truth-judge-llama2-7B} and informativeness\footnote{https://huggingface.co/allenai/truthfulqa-info-judge-llama2-7B}. \toxi is evaluated with a \textsc{HateBERT} classifier\footnote{https://huggingface.co/tomh/toxigen\_hatebert} to measure toxicity. Dataset sizes are detailed in \autoref{tab:datasize} (Appendix~\ref{sec:appendix-data_count}).

\paragraph{Contrastive Pairs Construction} For multiple-choice tasks, we generate positive prompts by concatenating questions with correct answers and generate negative prompts using a randomly chosen incorrect answer. For \trivia, a unique approach involves using a blank space as the incorrect answer. In \truthful, we utilize data from its multiple-choice format to identify crucial elements. For the \toxi task, we leverage the \texttt{RealToxicityPrompts} dataset~\citep{realtoxicityprompts}, selecting entries with a toxicity score exceeding 0.955 to serve as negative prompts. This helps us pinpoint elements contributing to toxic outputs.

\paragraph{Target LLMs} We evaluate the performance of \sadi\ in enhancing the baseline model (\baseline) - an instruction-tuned LLM, \llama \citep{llama}. To verify the generalizability of \sadi across various model backbones, we include three additional LLMs: \bloomz~\citep{BLOOMZ}, \mistral~\citep{Mistral}, \falcon~\citep{falcon}. These models are selected based on their demonstrated efficacy across diverse linguistic tasks and their widespread use in the research community. Furthermore, we extend our experiments to other models within the \textsc{BLOOMZ} family, specifically \bloomzfive, \bloomzone, and \bloomzthree, exploring how \sadi performs across different model sizes.

\subsection{Experimental Comparisons}
\label{sec:experimental-comparisons}
In addition to evaluating \sadi, we compare it against several approaches:

\textbf{Supervised fine-tuning (\sft)} We finetune all model parameters using the training dataset for each task, as previous works suggest that this approach serves as an upper bound for supervised finetuning. Specifically, we employ the AdamW optimizer with a learning rate of $2 \times 10^{-6}$ and a batch size of 4, conducting the fine-tuning across three epochs on four NVIDIA A-100 GPUs (80G).

\textbf{Inference-Time Intervention (\iti)} We follow~\citet{iti} in using contrastive pairs to identify the top heads for intervention. We sweep the hyperparameters of the heads involved and the strength of intervention to optimize results.

\textbf{Contrastive Activation Addition (\caa)} \citet{steer-vector} use the mean difference in the model’s activations at the position of the answer letter between all the positive and negative prompts to construct a fixed steering vector to shift activations.

\textbf{Our Approach (\sadi)} We compare three different configurations of the \sadi shift. \sadihidden applies \sadi to the identified key hidden states across all layers. \sadihead modifies activations from the outputs of all attention heads across all layers. \sadineuron is based on the outputs from each non-linear activation function in the FFN blocks across all layers. 

\subsection{Experimental Results}
\label{sec:main-results}

\begin{table}[t] \small
\centering
\caption{\label{overall-results-choice} The overall results of seven multiple-choice tasks in a zero-shot setting, performed by \llama. ``\sft + \sadi'' indicates that \sadi is applied to instruction fine-tuned models. A dash indicates that the training dataset is unavailable.}
\scalebox{0.9}{
\begin{tabular}{lcccccccc}
\toprule
\textbf{Task} & \textbf{\copa}  & \textbf{\story} & \textbf{\nli}    & \textbf{\mmlu}   & \textbf{\ssttwo}  & \textbf{\boolq}  & \textbf{\wino} & \textbf{\texttt{AVG}} \\ \hline
\baseline & 70.80 & 65.06 & 63.11  & 44.90  & 88.63 & 70.52  & 50.91  & 64.85 \\ 
\iti & 77.20 & 68.50 & 63.97 & 46.07 & 91.38 & 74.10 & 52.80 & 67.72 \\
\caa & 75.20 & 74.65 & 64.13 & 46.17 & 91.16 & 74.98 & 52.64 & 68.42 \\
\rowcolor[gray]{.93} \textbf{\sadi} & & &&& & &   &   \\
├ \sadihidden  & 81.00 & 55.99 & 59.28 & 45.66 & 92.15 & \textbf{76.25} & 52.64 & 66.14 \\
├ \sadineuron  & \textbf{82.20}   & 67.57   & 62.97 & 46.91 & 88.69    & 70.40 & 51.93 & 67.24 \\
└ \sadihead    & 78.80   & \textbf{79.75}  & \textbf{64.21} & \textbf{48.23} & \textbf{92.20}    & 74.35 & \textbf{53.04}  & \textbf{70.08} \\ \cdashline{1-9}[1pt/1pt]
\sft      & 93.20 & 96.49 & 90.07 & - & 96.70 & 88.75  &   78.37  & 90.59  \\
\rowcolor[gray]{.93} \textbf{\sft + \sadi} & & &&& & &    &  \\
├ \sadihidden & \textbf{94.90} & 96.49 & 90.07 & - & 96.76 & 88.91 & 78.37  & 90.91  \\
├ \sadineuron & 94.80 & 96.55 & \textbf{90.36} & - & \textbf{96.92} & 88.45 & 78.45  & 90.92   \\
└ \sadihead   & 94.60 & \textbf{96.55} &  90.30 & - & 96.81 & \textbf{88.94} &  \textbf{78.61}  & \textbf{90.97}  \\
\bottomrule
\end{tabular}
}
\end{table}

\paragraph{\sadi Significantly improves multiple-choice task performance.} As illustrated in \autoref{overall-results-choice}, \sadi demonstrates superior performance compared to the \baseline and other intervention methods across multiple-choice tasks. While \sft consistently outperforms other methods in tasks with available training data, such as \copa and \story, its high data resource demands limit its effectiveness in tasks with scarce data, like \mmlu. Both \iti and \caa show notable improvements, highlighting the effectiveness of intervention-based methods. \sadi, leveraging dynamic interventions, achieves optimal performance overall, significantly outperforming the \baseline. In comparison to fixed vector-based intervention methods (\iti and \caa), our dynamic intervention \sadi yields substantial gains across all tasks. Specifically, \sadi exceeds \iti and \caa by margins of +11.25 and +5.10, respectively, and surpasses the \baseline by a large margin of +14.69 in the \story task. Furthermore, \sadi can enhance the performance of task-specific fine-tuned models (\sft+\sadi), underscoring its practicality for precise and targeted interventions.

\begin{wraptable}[11]{t}{8cm} \small
\centering
\setlength{\tabcolsep}{3.5pt}
\caption{\label{mmlu-four} Average weighted accuracy on all four broad disciplines for \mmlu task with \llama.}
\begin{tabular}{lcccc}
\toprule
\textbf{Domain} & \textbf{Humanities} & \textbf{STEM} & \textbf{Social} & \textbf{Other} \\ \hline
\baseline & 46.68   & 34.26  & 54.62 & 49.71   \\
\rowcolor[gray]{.93} \textbf{\sadi} & & &&     \\
├ \sadihidden  & 48.99 &	35.03 &	55.82 &	48.51  \\
├ \sadineuron & 48.83   & 37.46  & 55.74 & 50.65  \\
└ \sadihead    & \textbf{49.76}   & \textbf{39.35}  & \textbf{56.21} & \textbf{52.31 }  \\
\bottomrule
\end{tabular}
\end{wraptable}

\paragraph{Improvement varies across \sadi configurations.} Performance gains from the three \sadi configurations vary, but all outperform \iti and \caa across tasks. Notably, \sadihead achieves the most significant improvements, enhancing average accuracy by up to +5.23. While \sadihidden and \sadineuron also demonstrate strong performance improvements in certain scenarios, such as 76.25 for \boolq and 82.20 for \copa, they occasionally show slight decreases in specific tasks, like \nli. Nevertheless, manipulating attention heads consistently results in improvements across all tasks and achieves the highest scores in most cases. For a detailed analysis, we present the results covering various domains of knowledge of \mmlu tasks, including humanities, STEM and social sciences and other are shown in \autoref{mmlu-four}. \sadi consistently enhances performances across these domains compared to the \baseline, with \sadihead yielding the highest improvements, up to +5.09 in the STEM domain.

\begin{table*}[t] \small
\centering
\caption{\label{overall-results-generation} The overall results of three open-ended generation tasks performed by \llama. The results of \trivia are obtained in a zero-shot setting, and we use 5-shot in-context learning in the \toxi and \truthful tasks. $^{\dagger}$ denotes results reproduced from other authors. }
\scalebox{0.9}{
\begin{tabular}{lcccccccc}
\toprule
\textbf{Task}     & \textbf{\trivia}  & \textbf{\toxi}    & \multicolumn{6}{c}{\textbf{\truthful}}\\ \cmidrule(rl){4-9}
\textbf{Metric}   & \textbf{EM} & \textbf{toxicity} $ \downarrow $     & \textbf{True}   & \textbf{Info}   & \textbf{True$\times$Info}  & \textbf{MC1}   & \textbf{MC2}   & \textbf{MC3}   \\ \hline
\baseline & 41.60 & 49.71 & 66.83   & 99.51  & 66.50   & 33.41  & 51.07  & 24.76  \\
\sft  & - & - & -& - & - & 24.20$^{\dagger}$ & - & -      \\
ITI & 42.80 & 45.27 & - & - & - & 34.64$^{\dagger}$ & 51.55$^{\dagger}$ & 25.32$^{\dagger}$     \\
CAA  & 43.20 & 49.71 & 71.60 & 83.84 & 60.03 & 34.03 & 52.76 &  25.62 \\
\rowcolor[gray]{.93} \textbf{\sadi} & & &  &&& & &      \\
├ \sadihidden  & 43.80 & 34.43 & 35.13 & 51.73 & 25.38 & 67.07 & 92.90 & 62.31      \\
├ \sadineuron & 43.50 & \textbf{17.14} & 74.54 & 93.51 & 69.71  & 34.88 & 52.50 & 25.79 \\
└ \sadihead   & \textbf{44.00} & 34.50 & \textbf{77.72} & 98.53  & \textbf{76.58} & \textbf{35.90} & \textbf{54.65} & \textbf{26.99} \\
\bottomrule
\end{tabular}
}
\end{table*}

\paragraph{\sadi improves open-ended generation task performance.} We further evaluate the performance of \sadi on the open-ended generation tasks in \autoref{overall-results-generation}. \sadi, in its three configurations, generally outperforms the \baseline, except for \sadihidden in the \truthful generation track. \sadihidden underperforms in the generation track, but shows significant improvements in the truthful multiple-choice track. This suggests that hidden states may be particularly sensitive to multiple-choice formats. Conversely, \sadihead significantly boosts truthfulness, with improvements reaching up to +10.08 on the True$\times$Info metric for \truthful. This underscores the generalizability of \sadi’s dynamic intervention, effectively tailoring activations to the semantics required by each task.

\begin{figure}[h]
    \centering
    \includegraphics[scale=0.5]{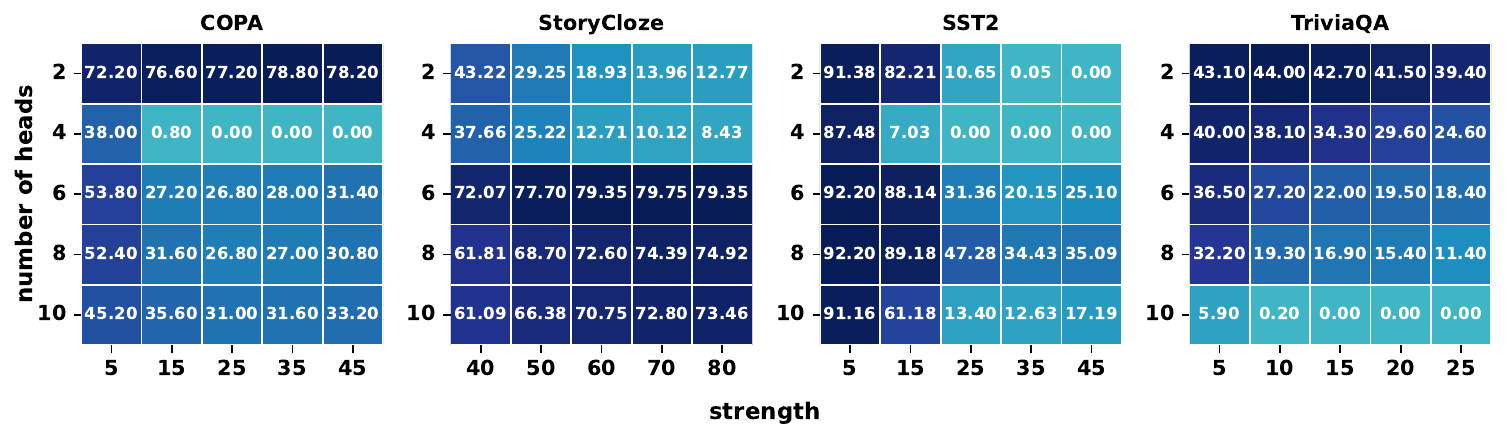}
    \caption{Results with varying intervention strength and numbers of key attention heads based on \copa, \story, \ssttwo, \trivia tasks with \llama.}
    \label{head-strength}
\end{figure}

\paragraph{Hyperparameters are task-specific.} In \autoref{head-strength}, we sweep two hyperparameters to control the intervention: the number of identified key attention heads, and the strength of intervention. Results indicate that optimal settings for these hyperparameters markedly vary across different tasks. This variability underscores the importance of carefully balancing the number of heads engaged and the scale of their amplification. For precise task performance optimization, it is recommended to search optimal hyperparameters using data from the validation sets specific to each task.

\subsection{Scrutinize Effects of Dynamic Intervention of \sadi}
\label{sec:scrutinize}

\paragraph{\sadi outperforms fixed steering and random intervention.}In \autoref{tab:scrutinize}, we conduct an ablation study to examine the contribution of Binary Masking (Step 2 in \autoref{sec:identify}) and Adaptive Steering (Step 3 in \autoref{sec:intervene}) in \sadi applied to tasks \copa, \story, \wino, \ssttwo, and \trivia. \sadi involves constructing an identification mask $M$, where we randomly assign K elements to 1 with the values of remainder elements set to 0 (termed ``random identify'' in \autoref{tab:scrutinize}). It can be observed from \autoref{tab:scrutinize} that random element identification leads to a notable performance decrease, with reductions as great as -7.88 (dropping from 79.75 to 71.87 on \story). Subsequently, the mask $M$ is applied to the activations corresponding to the semantics of user inputs in \sadi. 
\begin{wraptable}[12]{t}{6cm}\small
\centering
\caption{\label{tab:scrutinize} Ablation study for randomly identifying key elements and  intervening with a fixed steering vector for \sadihead with \llama.}
\setlength{\tabcolsep}{3.5pt}
\begin{tabular}{lccc}
\toprule
& \textbf{\sadi} & \begin{tabular}[c]{@{}l@{}}\textbf{random} \\ \textbf{identify}\end{tabular} & \begin{tabular}[c]{@{}l@{}}\textbf{fixed} \\ \textbf{steering} \end{tabular}  \\
\hline
\copa & \textbf{78.80} & 72.40  & 71.20  \\
\story & \textbf{79.75} & 71.87  & 69.42  \\
\wino &  \textbf{53.04}	& 52.88	& 52.56 \\
\ssttwo & \textbf{92.20} & 91.48 & 91.43 \\
\trivia & \textbf{44.00} & 42.90 & 41.50 \\
\bottomrule
\end{tabular}
\end{wraptable}
Additionally, we explore the effects of fixed steering, in which the mask $M$ is directly applied to the mean difference $D$, derived from the activations of contrastive pairs (see Eq. \ref{eq:difference}). If Step 3 (Eq. \ref{eq:intervene}) employs fixed steering, it updates activations as:
\begin{align}
    \bm{\mathcal{A}'}_{q} = \bm{\mathcal{A}}_{q} + \delta (\bm{D} \odot M) .
    \label{eq:intervene2}
\end{align}
The results in \autoref{tab:scrutinize} show that using a fixed steering vector leads to significant performance degradation compared to the semantic-adaptive approach of \sadi. This decline likely stems from a misalignment between the direction of intervention and that of input semantics.

\section{Discussion}
\label{sec:discussion}

In this section, we examine the generalizability across multiple LLMs (\autoref{sec:generalizability-llms}), different model sizes (\autoref{sec:generalizability-sizes}), few-shot settings (\autoref{sec:robustness-fewshots}), and multilingual scenarios (\autoref{sec:robustness-multilingual}).

\subsection{Generalizability across Multiple LLMs}
\label{sec:generalizability-llms}
\begin{table}[h] \scriptsize
\centering
\setlength{\tabcolsep}{3.5pt}
\caption{\label{3LLMs} Generalizability evaluation of \sadi given by \bloomz, \mistral, and \falcon on the \copa, \boolq and \nli tasks .}
\begin{tabular}{lccccccccc}
\toprule
\textbf{Task }    & \multicolumn{3}{c}{\textbf{\copa}} & \multicolumn{3}{c}{\textbf{\boolq}}& \multicolumn{3}{c}{\textbf{\nli}}  \\ \cmidrule(rl){2-4} \cmidrule(rl){5-7} \cmidrule(rl){8-10}
\textbf{LLMs}      & \textbf{\textsc{BLOOMZ}}  & \textbf{\textsc{Mistral}} & \textbf{\textsc{Falcon}}  & \textbf{\textsc{BLOOMZ}}  & \textbf{\textsc{Mistral}} & \textbf{\textsc{Falcon}}    & \textbf{\textsc{BLOOMZ}}  &\textbf{ \textsc{Mistral}} & \textbf{\textsc{Falcon}}   \\ \hline
\baseline  & 76.40   & 84.80   & 62.20   & 91.28   & 71.80   & 71.83   & 54.81   & 53.33   & 52.29   \\
\sft& 86.80   & 86.80   & 88.20   & 90.52   & 86.57   & 84.22   & 57.41   & 89.94   & 55.45   \\
\rowcolor[gray]{.93} \textbf{\sadi} &  &  &  &  &  &  &  &  &  \\
├ \sadihidden   & 76.40   & \textbf{93.00} & 60.60   & 91.31 & 49.43   & 62.73   &  53.25 & 35.69 & 43.67 \\
├ \sadineuron   & \textbf{82.20} & 84.20 & \textbf{62.40} & \textbf{91.40} & \textbf{74.61} & \textbf{74.48} & \textbf{56.43} & \textbf{54.07} & 52.48 \\
└ \sadihead     & \textbf{82.20} & 92.00 & 62.20   & 91.28   & 67.38   & 73.23 & 55.37 & 53.65 & \textbf{54.83} \\
\bottomrule
\end{tabular}
\end{table}
\paragraph{\sadi enhances performance across multiple LLMs in various tasks.} An important question is whether \sadi can generalize over various LLMs. We apply \sadi to three other well-performing LLMs: \bloomz, \mistral, and \falcon in \copa, \boolq and \nli tasks. According to the results in \autoref{3LLMs}, \sadi consistently enhances performance across tasks compared to the \baseline, despite varying improvement levels by configuration. It is noteworthy that the \sadineuron configuration with \bloomz achieves the most substantial performance gains in all three tasks, demonstrating that different models may exhibit distinct functional elements.

\subsection{Generalizability across Model Sizes}
\label{sec:generalizability-sizes}

\begin{wraptable}[8]{r}{6cm} \small
    \centering
    \setlength{\tabcolsep}{3.5pt}
    \caption{\label{scale}Generalizability evaluation on \textsc{bloomz} series in \copa task.}
    \begin{tabular}{lcccc}
    \toprule
\textbf{Size} & \textbf{7b} &\textbf{ 3b} & \textbf{1.1b} & \textbf{560m} \\ \hline
\baseline  & 70.8 & 79.2 & 49.8 & 50.0 \\
\sft & \textbf{88.8} & \textbf{85.8} & 50.6 & 52.0 \\
\rowcolor[gray]{.93} \textbf{\sadi} &  &  &  & \\
├ \sadihidden & 76.4 & 79.2 & 50.0 & 50.0 \\
├ \sadineuron  & 74.0 & 79.2 & 50.0 & 52.6 \\
└ \sadihead  & 78.8 & 79.2 & \textbf{51.8} & \textbf{54.4} \\
\bottomrule
\end{tabular}
\end{wraptable}
\paragraph{\sadi outperforms \sft in smaller LLMs.} We further investigate the effectiveness of \sadi on the \textsc{bloomz} series across various model sizes. As shown in \autoref{scale}, \sadi maintains its improvement over \baseline even with smaller LLM sizes, confirming its generalizability across model sizes. Notably, \sadi outperforms \sft with incremental gains of up to +1.2 for \bloomzone and +2.4 for \bloomzfive, highlighting its effectiveness, especially in smaller models.

\subsection{Generalizability in Few-Shot Settings}
\label{sec:robustness-fewshots}

\paragraph{\sadi improves few-shot performance but with less gains.} In \autoref{few-shots}, we compare \sadihead to the \baseline across zero-shot and few-shot settings on the \sstfive, \wino, and \truthful tasks. The results highlight the generalizability of \sadi in enhancing model performance with few-shot prompting across various tasks. While manipulating heads improves the performances in few-shot settings, the gains are less pronounced compared to those in zero-shot settings. This suggests that few-shot examples already provide a strong learning signal (to both \baseline and \sadi), somewhat overshadowing the additional benefits derived from head manipulation.

\begin{table}[h] \small
\centering
\setlength{\tabcolsep}{3.5pt}
\caption{\label{few-shots} Comparisons between few-shot and zero-shot on the \sstfive, \wino, and \truthful.}
\begin{tabular}{lcccccccccc}
\toprule
\textbf{Task} & \multicolumn{2}{c}{\textbf{\sstfive}} & \multicolumn{2}{c}{\textbf{\wino}} & \multicolumn{2}{c}{\textbf{\truthful MC1}} & \multicolumn{2}{c}{\textbf{\truthful MC2}} & \multicolumn{2}{l}{\textbf{\truthful MC3}} \\ \cmidrule(rl){2-3} \cmidrule(rl){4-5} \cmidrule(rl){6-7} \cmidrule(rl){8-9} \cmidrule(rl){10-11}
\textbf{Configuration }        &\textbf{ 0-shot}     & \textbf{5-shot}     & \textbf{0-shot}       & \textbf{5-shot }       & \textbf{0-shot}        & \textbf{5-shot}          & \textbf{0-shot}          & \textbf{5-shot}          & \textbf{0-shot}          & \textbf{5-shot }         \\ \hline
\baseline & 28.24      & 53.21       & 50.91         & 52.01          & 27.66           & 33.41            & 44.45           & 51.07            & 20.59           & 24.76            \\
\sadihead    & 35.43      & 54.07       & 53.04         & 53.35          & 32.19           & 35.99            & 50.81           & 54.65            & 24.83           & 26.99           \\
\bottomrule
\end{tabular}
\end{table}

\subsection{Generalizability in Multilingual Scenarios}
\label{sec:robustness-multilingual}
\begin{table}[h] \small
\centering
\setlength{\tabcolsep}{5.5pt}
\caption{\label{multilingual} Evaluating \sadi on multilingual task \xcopa with \llama.}
\begin{tabular}{lccccccccc}
\toprule
\textbf{Language} & \textbf{id }   & \textbf{it}    & \textbf{sw }   &\textbf{ ta}    & \textbf{th}    & \textbf{tr}    & \textbf{vi}    & \textbf{zh}   & \textbf{AVG} \\ \hline
\baseline & 51.40 & 61.20 & 50.20 & 49.40 & 50.80 & 49.40 & 51.80 & 62.80  & 53.38\\
\rowcolor[gray]{.93} \textbf{\sadi} & & &&   & & &&  &   \\
├ \sadihidden & 51.40 & 62.40 & 50.00 & \textbf{50.00} & 51.20 & 48.80 & 52.20 & 64.80 & 53.85 \\
├ \sadineuron & \textbf{63.60} & 68.80 & 50.20 & 48.80 & \textbf{53.80} & 50.60 & \textbf{60.40} & \textbf{70.40} & 58.33 \\
└ \sadihead   & 62.60 & \textbf{70.60} & \textbf{50.80} & 49.60 & 51.40 & \textbf{51.60} & 60.20 & 70.10 & \textbf{58.36} \\
\bottomrule
\end{tabular}
\end{table}

\paragraph{\sadi enhances performance in multilingual scenarios.} Although our primary experiments are in English, extending \sadi to multilingual scenarios reveals its broader applicability. We further evaluate \sadi on the multilingual \xcopa task \citep{xcopa}, covering eight languages: Indonesian (id), Italian (it), Swahili (sw), Tamil (ta), Thai (th), Turkish (tr), Vietnamese (vi), Chinese (zh). \autoref{multilingual} illustrates varying degrees of performance enhancements across different languages. It can be observed that Indonesian shows the highest improvement, while Swahili gains the least. Despite these variations, \sadi consistently boosts performance across diverse language settings and configurations. A detailed analysis of the key components effective language-wise is provided in Appendix~\ref{sec:overlap}. Additional insights from cross-lingual evaluations are discussed in Appendix~\ref{appendix-crosslingual}.

\section{Analysis}
\label{sec:analysis}
In this section, we analyze activation difference distribution patterns for key model elements (\autoref{sec:analysis-difference}), and how \sadi behaves under varying numbers of contrastive pairs (\autoref{sec:analysis-data}).

\subsection{Characteristics of Activation Difference}
\label{sec:analysis-difference}

\begin{figure}[h]
\begin{minipage}{0.67\linewidth}
\centering
\subfigure[Difference for all heads across layers. Darker blue represents bigger difference.]{\label{cnt-head}\includegraphics[scale=0.45]{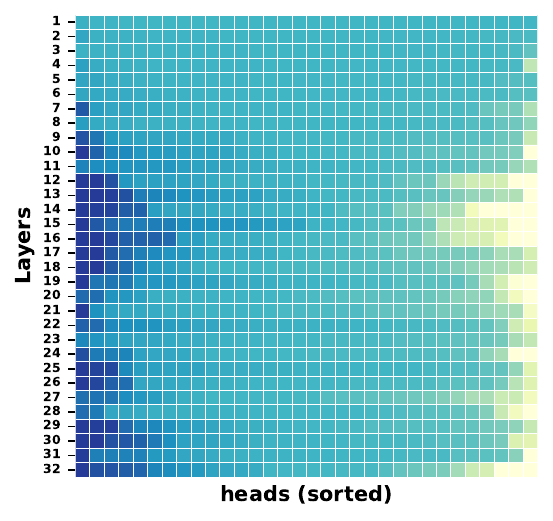}} 
\hspace{1em}
\subfigure[Distribution of top-100 activation difference of neurons and hidden states. The y-axis represents the number of neurons or hidden states in top-100.]{\label{cnt-neuron}\includegraphics[scale=0.4]{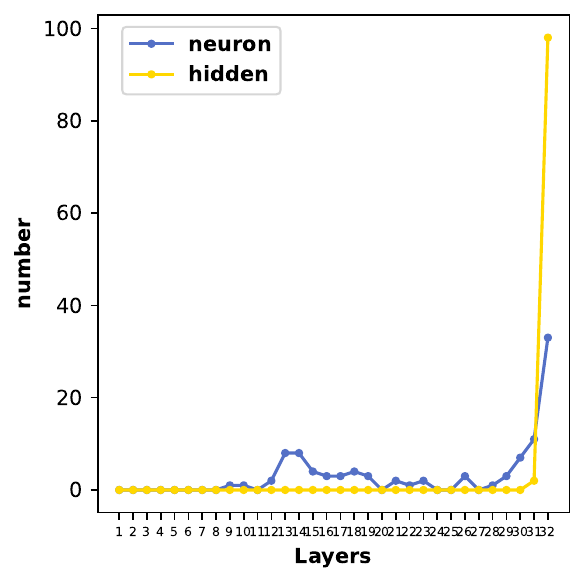}} 
\caption{\label{cnt-distribution} Activation difference of each head across layers and the distribution of top-100 activation difference of neurons and hidden states with \llama in \story.}
\end{minipage}
\begin{minipage}{0.33\linewidth}
\centering
\includegraphics[width=0.99\columnwidth]{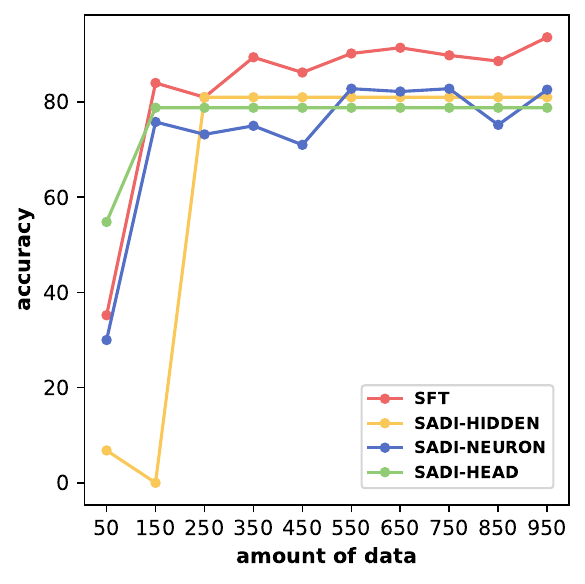}
\caption{\centering Relationship between accuracy and the amount of contrastive pairs.}
\label{amount-contrsative}
\end{minipage}
\end{figure}

Figure \autoref{cnt-head} reveals activation patterns of attention heads in the middle to latter layers for contrastive pairs in \story, indicating primary information processing within these layers. 
Figure \autoref{cnt-neuron} shows that activation differences in neuron activity and hidden states are concentrated in the latter layers, with the most significant discrepancies observed in the final layer. These consistent patterns across tasks (see Appendix~\ref{sec:appendix-distribution}) lend support to the functional segregation hypothesis, which posits 
that latter layers are associated with language generation and middle layers are responsible for reasoning \citep{handle}. Given this, our interventions on attention heads likely influence both reasoning and generation, contributing to the consistent improvements.

\subsection{\sadi and \sft with varying data}
\label{sec:analysis-data}

We assess the impact of varying amounts of contrastive pairs on \sadi and \sft in \copa task. As shown in \autoref{amount-contrsative}, \sft performance improves with an increasing number of fine-tuning data. In contrast, \sadi achieves optimal results with significantly fewer pairs, e.g., only 150 items are sufficient to calculate an identification mask for targeting critical heads for intervention. This highlights \sadi's effectiveness and efficiency in low-resource conditions.

\section{Conclusion}

In this study, we propose Semantics-Adaptive Dynamic Intervention (\sadi), a novel approach designed to dynamically steer model behavior by adapting to the semantic contexts of inputs. \sadi enhances model adaptability by modulating the activations of the identified critical model elements during inference, taking into account the directions of input semantics. Our extensive experiments across various tasks, LLMs backbones, and languages settings have demonstrated that \sadi significantly outperforms established baselines, offering generalizable improvements without requiring additional training. Our study advances the field of ``activation engineering'' in LLMs, with the potential to inform the development of more advanced LLM intervention techniques.

\section*{Ethics Statement}
This work presents Semantics-Adaptive Dynamic Intervention (\sadi), a method designed to enhance the performance of large language models (LLMs) by dynamically adjusting their activations without additional training. We have conducted extensive experiments across diverse tasks to evaluate SADI's effectiveness, but we recognize that biases in the underlying models and datasets may still affect outcomes. We encourage practitioners to use SADI responsibly, with careful consideration of fairness, accountability, and transparency. No human subjects were involved in this research, and all experiments were conducted using publicly available models and datasets, adhering to their respective licenses and use policies.

\section*{Reproducibility Statement}
We are committed to ensuring the reproducibility of our findings in this work. To facilitate this, we provide comprehensive details of our proposed Semantics-Adaptive Dynamic Intervention (\sadi) method in \autoref{sec:method} of the main paper, including the algorithms and mechanisms for dynamic steering vector generation and application to different model components. The experimental setups, including model configurations, datasets, and evaluation metrics, are thoroughly described in \autoref{sec:experiments}. We utilize publicly available models: \llama, \bloomz, \mistral, and \falcon, and detail any modifications or specific settings used during experimentation. All datasets employed in our evaluation are standard benchmarks, and we include references and links to these resources for accessibility. To further support reproducibility, we will release our code used for experiments upon publication, enabling other researchers to replicate our results and extend our work.

\bibliography{iclr2025_conference}

\begin{thebibliography}{40}
\providecommand{\natexlab}[1]{#1}
\providecommand{\url}[1]{\texttt{#1}}
\expandafter\ifx\csname urlstyle\endcsname\relax
  \providecommand{\doi}[1]{doi: #1}\else
  \providecommand{\doi}{doi: \begingroup \urlstyle{rm}\Url}\fi

\bibitem[Almazrouei et~al.(2023)Almazrouei, Alobeidli, Alshamsi, Cappelli, Cojocaru, Debbah, Goffinet, Hesslow, Launay, Malartic, Mazzotta, Noune, Pannier, and Penedo]{falcon}
Ebtesam Almazrouei, Hamza Alobeidli, Abdulaziz Alshamsi, Alessandro Cappelli, Ruxandra Cojocaru, M{\'{e}}rouane Debbah, {\'{E}}tienne Goffinet, Daniel Hesslow, Julien Launay, Quentin Malartic, Daniele Mazzotta, Badreddine Noune, Baptiste Pannier, and Guilherme Penedo.
\newblock The falcon series of open language models.
\newblock \emph{CoRR}, abs/2311.16867, 2023.
\newblock \doi{10.48550/ARXIV.2311.16867}.
\newblock URL \url{https://doi.org/10.48550/arXiv.2311.16867}.

\bibitem[Anil et~al.(2023{\natexlab{a}})Anil, Borgeaud, Wu, Alayrac, Yu, Soricut, Schalkwyk, Dai, Hauth, Millican, Silver, Petrov, Johnson, Antonoglou, Schrittwieser, Glaese, Chen, Pitler, Lillicrap, Lazaridou, Firat, Molloy, Isard, Barham, Hennigan, Lee, Viola, Reynolds, Xu, Doherty, Collins, Meyer, Rutherford, Moreira, Ayoub, Goel, Tucker, Piqueras, Krikun, Barr, Savinov, Danihelka, Roelofs, White, Andreassen, von Glehn, Yagati, Kazemi, Gonzalez, Khalman, Sygnowski, and et~al.]{gemini}
Rohan Anil, Sebastian Borgeaud, Yonghui Wu, Jean{-}Baptiste Alayrac, Jiahui Yu, Radu Soricut, Johan Schalkwyk, Andrew~M. Dai, Anja Hauth, Katie Millican, David Silver, Slav Petrov, Melvin Johnson, Ioannis Antonoglou, Julian Schrittwieser, Amelia Glaese, Jilin Chen, Emily Pitler, Timothy~P. Lillicrap, Angeliki Lazaridou, Orhan Firat, James Molloy, Michael Isard, Paul~Ronald Barham, Tom Hennigan, Benjamin Lee, Fabio Viola, Malcolm Reynolds, Yuanzhong Xu, Ryan Doherty, Eli Collins, Clemens Meyer, Eliza Rutherford, Erica Moreira, Kareem Ayoub, Megha Goel, George Tucker, Enrique Piqueras, Maxim Krikun, Iain Barr, Nikolay Savinov, Ivo Danihelka, Becca Roelofs, Ana{\"{\i}}s White, Anders Andreassen, Tamara von Glehn, Lakshman Yagati, Mehran Kazemi, Lucas Gonzalez, Misha Khalman, Jakub Sygnowski, and et~al.
\newblock Gemini: {A} family of highly capable multimodal models.
\newblock \emph{CoRR}, abs/2312.11805, 2023{\natexlab{a}}.
\newblock \doi{10.48550/ARXIV.2312.11805}.
\newblock URL \url{https://doi.org/10.48550/arXiv.2312.11805}.

\bibitem[Anil et~al.(2023{\natexlab{b}})Anil, Dai, Firat, Johnson, Lepikhin, Passos, Shakeri, Taropa, Bailey, Chen, Chu, Clark, Shafey, Huang, Meier{-}Hellstern, Mishra, Moreira, Omernick, Robinson, Ruder, Tay, Xiao, Xu, Zhang, {\'{A}}brego, Ahn, Austin, Barham, Botha, Bradbury, Brahma, Brooks, Catasta, Cheng, Cherry, Choquette{-}Choo, Chowdhery, Crepy, Dave, Dehghani, Dev, Devlin, D{\'{\i}}az, Du, Dyer, Feinberg, Feng, Fienber, Freitag, Garcia, Gehrmann, Gonzalez, and et~al.]{palm}
Rohan Anil, Andrew~M. Dai, Orhan Firat, Melvin Johnson, Dmitry Lepikhin, Alexandre Passos, Siamak Shakeri, Emanuel Taropa, Paige Bailey, Zhifeng Chen, Eric Chu, Jonathan~H. Clark, Laurent~El Shafey, Yanping Huang, Kathy Meier{-}Hellstern, Gaurav Mishra, Erica Moreira, Mark Omernick, Kevin Robinson, Sebastian Ruder, Yi~Tay, Kefan Xiao, Yuanzhong Xu, Yujing Zhang, Gustavo~Hern{\'{a}}ndez {\'{A}}brego, Junwhan Ahn, Jacob Austin, Paul Barham, Jan~A. Botha, James Bradbury, Siddhartha Brahma, Kevin Brooks, Michele Catasta, Yong Cheng, Colin Cherry, Christopher~A. Choquette{-}Choo, Aakanksha Chowdhery, Cl{\'{e}}ment Crepy, Shachi Dave, Mostafa Dehghani, Sunipa Dev, Jacob Devlin, Mark D{\'{\i}}az, Nan Du, Ethan Dyer, Vladimir Feinberg, Fangxiaoyu Feng, Vlad Fienber, Markus Freitag, Xavier Garcia, Sebastian Gehrmann, Lucas Gonzalez, and et~al.
\newblock Palm 2 technical report.
\newblock \emph{CoRR}, abs/2305.10403, 2023{\natexlab{b}}.
\newblock \doi{10.48550/ARXIV.2305.10403}.
\newblock URL \url{https://doi.org/10.48550/arXiv.2305.10403}.

\bibitem[Bai et~al.(2022)Bai, Jones, Ndousse, Askell, Chen, DasSarma, Drain, Fort, Ganguli, Henighan, Joseph, Kadavath, Kernion, Conerly, Showk, Elhage, Hatfield{-}Dodds, Hernandez, Hume, Johnston, Kravec, Lovitt, Nanda, Olsson, Amodei, Brown, Clark, McCandlish, Olah, Mann, and Kaplan]{rhlf}
Yuntao Bai, Andy Jones, Kamal Ndousse, Amanda Askell, Anna Chen, Nova DasSarma, Dawn Drain, Stanislav Fort, Deep Ganguli, Tom Henighan, Nicholas Joseph, Saurav Kadavath, Jackson Kernion, Tom Conerly, Sheer~El Showk, Nelson Elhage, Zac Hatfield{-}Dodds, Danny Hernandez, Tristan Hume, Scott Johnston, Shauna Kravec, Liane Lovitt, Neel Nanda, Catherine Olsson, Dario Amodei, Tom~B. Brown, Jack Clark, Sam McCandlish, Chris Olah, Benjamin Mann, and Jared Kaplan.
\newblock Training a helpful and harmless assistant with reinforcement learning from human feedback.
\newblock \emph{CoRR}, abs/2204.05862, 2022.
\newblock \doi{10.48550/ARXIV.2204.05862}.
\newblock URL \url{https://doi.org/10.48550/arXiv.2204.05862}.

\bibitem[Bowman et~al.(2015)Bowman, Angeli, Potts, and Manning]{nli}
Samuel~R. Bowman, Gabor Angeli, Christopher Potts, and Christopher~D. Manning.
\newblock A large annotated corpus for learning natural language inference.
\newblock In Llu{\'{\i}}s M{\`{a}}rquez, Chris Callison{-}Burch, Jian Su, Daniele Pighin, and Yuval Marton (eds.), \emph{Proceedings of the 2015 Conference on Empirical Methods in Natural Language Processing, {EMNLP} 2015, Lisbon, Portugal, September 17-21, 2015}, pp.\  632--642. The Association for Computational Linguistics, 2015.
\newblock \doi{10.18653/V1/D15-1075}.
\newblock URL \url{https://doi.org/10.18653/v1/d15-1075}.

\bibitem[Chen et~al.(2024)Chen, Sun, Jiao, Lian, Kang, Wang, and Xu]{trfr}
Zhongzhi Chen, Xingwu Sun, Xianfeng Jiao, Fengzong Lian, Zhanhui Kang, Di~Wang, and Chengzhong Xu.
\newblock Truth forest: Toward multi-scale truthfulness in large language models through intervention without tuning.
\newblock In Michael~J. Wooldridge, Jennifer~G. Dy, and Sriraam Natarajan (eds.), \emph{Thirty-Eighth {AAAI} Conference on Artificial Intelligence, {AAAI} 2024, Thirty-Sixth Conference on Innovative Applications of Artificial Intelligence, {IAAI} 2024, Fourteenth Symposium on Educational Advances in Artificial Intelligence, {EAAI} 2014, February 20-27, 2024, Vancouver, Canada}, pp.\  20967--20974. {AAAI} Press, 2024.
\newblock \doi{10.1609/AAAI.V38I19.30087}.
\newblock URL \url{https://doi.org/10.1609/aaai.v38i19.30087}.

\bibitem[Clark et~al.(2019)Clark, Lee, Chang, Kwiatkowski, Collins, and Toutanova]{boolq}
Christopher Clark, Kenton Lee, Ming{-}Wei Chang, Tom Kwiatkowski, Michael Collins, and Kristina Toutanova.
\newblock Boolq: Exploring the surprising difficulty of natural yes/no questions.
\newblock In Jill Burstein, Christy Doran, and Thamar Solorio (eds.), \emph{Proceedings of the 2019 Conference of the North American Chapter of the Association for Computational Linguistics: Human Language Technologies, {NAACL-HLT} 2019, Minneapolis, MN, USA, June 2-7, 2019, Volume 1 (Long and Short Papers)}, pp.\  2924--2936. Association for Computational Linguistics, 2019.
\newblock \doi{10.18653/V1/N19-1300}.
\newblock URL \url{https://doi.org/10.18653/v1/n19-1300}.

\bibitem[Ding et~al.(2023)Ding, Chen, Xu, Qin, Hu, Liu, Sun, and Zhou]{instruction}
Ning Ding, Yulin Chen, Bokai Xu, Yujia Qin, Shengding Hu, Zhiyuan Liu, Maosong Sun, and Bowen Zhou.
\newblock Enhancing chat language models by scaling high-quality instructional conversations.
\newblock In Houda Bouamor, Juan Pino, and Kalika Bali (eds.), \emph{Proceedings of the 2023 Conference on Empirical Methods in Natural Language Processing, {EMNLP} 2023, Singapore, December 6-10, 2023}, pp.\  3029--3051. Association for Computational Linguistics, 2023.
\newblock \doi{10.18653/V1/2023.EMNLP-MAIN.183}.
\newblock URL \url{https://doi.org/10.18653/v1/2023.emnlp-main.183}.

\bibitem[Gehman et~al.(2020)Gehman, Gururangan, Sap, Choi, and Smith]{realtoxicityprompts}
Samuel Gehman, Suchin Gururangan, Maarten Sap, Yejin Choi, and Noah~A. Smith.
\newblock Realtoxicityprompts: Evaluating neural toxic degeneration in language models.
\newblock In Trevor Cohn, Yulan He, and Yang Liu (eds.), \emph{Findings of the Association for Computational Linguistics: {EMNLP} 2020, Online Event, 16-20 November 2020}, volume {EMNLP} 2020 of \emph{Findings of {ACL}}, pp.\  3356--3369. Association for Computational Linguistics, 2020.
\newblock \doi{10.18653/V1/2020.FINDINGS-EMNLP.301}.
\newblock URL \url{https://doi.org/10.18653/v1/2020.findings-emnlp.301}.

\bibitem[Gordon et~al.(2012)Gordon, Kozareva, and Roemmele]{copa}
Andrew Gordon, Zornitsa Kozareva, and Melissa Roemmele.
\newblock {S}em{E}val-2012 task 7: Choice of plausible alternatives: An evaluation of commonsense causal reasoning.
\newblock In Eneko Agirre, Johan Bos, Mona Diab, Suresh Manandhar, Yuval Marton, and Deniz Yuret (eds.), \emph{*{SEM} 2012: The First Joint Conference on Lexical and Computational Semantics {--} Volume 1: Proceedings of the main conference and the shared task, and Volume 2: Proceedings of the Sixth International Workshop on Semantic Evaluation ({S}em{E}val 2012)}, pp.\  394--398, Montr{\'e}al, Canada, 7-8 June 2012. Association for Computational Linguistics.
\newblock URL \url{https://aclanthology.org/S12-1052}.

\bibitem[Hartvigsen et~al.(2022)Hartvigsen, Gabriel, Palangi, Sap, Ray, and Kamar]{toxigen}
Thomas Hartvigsen, Saadia Gabriel, Hamid Palangi, Maarten Sap, Dipankar Ray, and Ece Kamar.
\newblock Toxigen: {A} large-scale machine-generated dataset for adversarial and implicit hate speech detection.
\newblock In Smaranda Muresan, Preslav Nakov, and Aline Villavicencio (eds.), \emph{Proceedings of the 60th Annual Meeting of the Association for Computational Linguistics (Volume 1: Long Papers), {ACL} 2022, Dublin, Ireland, May 22-27, 2022}, pp.\  3309--3326. Association for Computational Linguistics, 2022.
\newblock \doi{10.18653/V1/2022.ACL-LONG.234}.
\newblock URL \url{https://doi.org/10.18653/v1/2022.acl-long.234}.

\bibitem[Hendrycks et~al.(2021)Hendrycks, Burns, Basart, Zou, Mazeika, Song, and Steinhardt]{mmlu}
Dan Hendrycks, Collin Burns, Steven Basart, Andy Zou, Mantas Mazeika, Dawn Song, and Jacob Steinhardt.
\newblock Measuring massive multitask language understanding.
\newblock In \emph{9th International Conference on Learning Representations, {ICLR} 2021, Virtual Event, Austria, May 3-7, 2021}. OpenReview.net, 2021.
\newblock URL \url{https://openreview.net/forum?id=d7KBjmI3GmQ}.

\bibitem[Hernandez et~al.(2023)Hernandez, Li, and Andreas]{engineering2}
Evan Hernandez, Belinda~Z Li, and Jacob Andreas.
\newblock Inspecting and editing knowledge representations in language models.
\newblock \emph{arXiv preprint arXiv:2304.00740}, 2023.

\bibitem[Jiang et~al.(2023)Jiang, Sablayrolles, Mensch, Bamford, Chaplot, de~Las~Casas, Bressand, Lengyel, Lample, Saulnier, Lavaud, Lachaux, Stock, Scao, Lavril, Wang, Lacroix, and Sayed]{Mistral}
Albert~Q. Jiang, Alexandre Sablayrolles, Arthur Mensch, Chris Bamford, Devendra~Singh Chaplot, Diego de~Las~Casas, Florian Bressand, Gianna Lengyel, Guillaume Lample, Lucile Saulnier, L{\'{e}}lio~Renard Lavaud, Marie{-}Anne Lachaux, Pierre Stock, Teven~Le Scao, Thibaut Lavril, Thomas Wang, Timoth{\'{e}}e Lacroix, and William~El Sayed.
\newblock Mistral 7b.
\newblock \emph{CoRR}, abs/2310.06825, 2023.
\newblock \doi{10.48550/ARXIV.2310.06825}.
\newblock URL \url{https://doi.org/10.48550/arXiv.2310.06825}.

\bibitem[Joshi et~al.(2017)Joshi, Choi, Weld, and Zettlemoyer]{triviaqa}
Mandar Joshi, Eunsol Choi, Daniel Weld, and Luke Zettlemoyer.
\newblock {T}rivia{QA}: A large scale distantly supervised challenge dataset for reading comprehension.
\newblock In Regina Barzilay and Min-Yen Kan (eds.), \emph{Proceedings of the 55th Annual Meeting of the Association for Computational Linguistics (Volume 1: Long Papers)}, pp.\  1601--1611, Vancouver, Canada, July 2017. Association for Computational Linguistics.
\newblock \doi{10.18653/v1/P17-1147}.
\newblock URL \url{https://aclanthology.org/P17-1147}.

\bibitem[Li et~al.(2023{\natexlab{a}})Li, Wang, Zhang, Zhang, and Zong]{function-head}
Chong Li, Shaonan Wang, Yunhao Zhang, Jiajun Zhang, and Chengqing Zong.
\newblock Interpreting and exploiting functional specialization in multi-head attention under multi-task learning.
\newblock In Houda Bouamor, Juan Pino, and Kalika Bali (eds.), \emph{Proceedings of the 2023 Conference on Empirical Methods in Natural Language Processing, {EMNLP} 2023, Singapore, December 6-10, 2023}, pp.\  16460--16476. Association for Computational Linguistics, 2023{\natexlab{a}}.
\newblock \doi{10.18653/V1/2023.EMNLP-MAIN.1026}.
\newblock URL \url{https://doi.org/10.18653/v1/2023.emnlp-main.1026}.

\bibitem[Li et~al.(2023{\natexlab{b}})Li, Patel, Vi{\'{e}}gas, Pfister, and Wattenberg]{iti}
Kenneth Li, Oam Patel, Fernanda~B. Vi{\'{e}}gas, Hanspeter Pfister, and Martin Wattenberg.
\newblock Inference-time intervention: Eliciting truthful answers from a language model.
\newblock In Alice Oh, Tristan Naumann, Amir Globerson, Kate Saenko, Moritz Hardt, and Sergey Levine (eds.), \emph{Advances in Neural Information Processing Systems 36: Annual Conference on Neural Information Processing Systems 2023, NeurIPS 2023, New Orleans, LA, USA, December 10 - 16, 2023}, 2023{\natexlab{b}}.
\newblock URL \url{http://papers.nips.cc/paper\_files/paper/2023/hash/81b8390039b7302c909cb769f8b6cd93-Abstract-Conference.html}.

\bibitem[Lin et~al.(2022)Lin, Hilton, and Evans]{truthfulqa}
Stephanie Lin, Jacob Hilton, and Owain Evans.
\newblock Truthfulqa: Measuring how models mimic human falsehoods.
\newblock In Smaranda Muresan, Preslav Nakov, and Aline Villavicencio (eds.), \emph{Proceedings of the 60th Annual Meeting of the Association for Computational Linguistics (Volume 1: Long Papers), {ACL} 2022, Dublin, Ireland, May 22-27, 2022}, pp.\  3214--3252. Association for Computational Linguistics, 2022.
\newblock \doi{10.18653/V1/2022.ACL-LONG.229}.
\newblock URL \url{https://doi.org/10.18653/v1/2022.acl-long.229}.

\bibitem[Liu et~al.(2023)Liu, Xing, and Zou]{context-vector}
Sheng Liu, Lei Xing, and James Zou.
\newblock In-context vectors: Making in context learning more effective and controllable through latent space steering.
\newblock \emph{CoRR}, abs/2311.06668, 2023.
\newblock \doi{10.48550/ARXIV.2311.06668}.
\newblock URL \url{https://doi.org/10.48550/arXiv.2311.06668}.

\bibitem[Longpre et~al.(2023)Longpre, Hou, Vu, Webson, Chung, Tay, Zhou, Le, Zoph, Wei, and Roberts]{flan}
Shayne Longpre, Le~Hou, Tu~Vu, Albert Webson, Hyung~Won Chung, Yi~Tay, Denny Zhou, Quoc~V. Le, Barret Zoph, Jason Wei, and Adam Roberts.
\newblock The flan collection: Designing data and methods for effective instruction tuning.
\newblock In Andreas Krause, Emma Brunskill, Kyunghyun Cho, Barbara Engelhardt, Sivan Sabato, and Jonathan Scarlett (eds.), \emph{International Conference on Machine Learning, {ICML} 2023, 23-29 July 2023, Honolulu, Hawaii, {USA}}, volume 202 of \emph{Proceedings of Machine Learning Research}, pp.\  22631--22648. {PMLR}, 2023.
\newblock URL \url{https://proceedings.mlr.press/v202/longpre23a.html}.

\bibitem[Mesnard et~al.(2024)Mesnard, Hardin, Dadashi, Bhupatiraju, Pathak, Sifre, Rivi{\`{e}}re, Kale, Love, Tafti, Hussenot, Chowdhery, Roberts, Barua, Botev, Castro{-}Ros, Slone, H{\'{e}}liou, Tacchetti, Bulanova, Paterson, Tsai, Shahriari, Lan, Choquette{-}Choo, Crepy, Cer, Ippolito, Reid, Buchatskaya, Ni, Noland, Yan, Tucker, Muraru, Rozhdestvenskiy, Michalewski, Tenney, Grishchenko, Austin, Keeling, Labanowski, Lespiau, Stanway, Brennan, Chen, Ferret, Chiu, and et~al.]{gemma}
Thomas Mesnard, Cassidy Hardin, Robert Dadashi, Surya Bhupatiraju, Shreya Pathak, Laurent Sifre, Morgane Rivi{\`{e}}re, Mihir~Sanjay Kale, Juliette Love, Pouya Tafti, L{\'{e}}onard Hussenot, Aakanksha Chowdhery, Adam Roberts, Aditya Barua, Alex Botev, Alex Castro{-}Ros, Ambrose Slone, Am{\'{e}}lie H{\'{e}}liou, Andrea Tacchetti, Anna Bulanova, Antonia Paterson, Beth Tsai, Bobak Shahriari, Charline~Le Lan, Christopher~A. Choquette{-}Choo, Cl{\'{e}}ment Crepy, Daniel Cer, Daphne Ippolito, David Reid, Elena Buchatskaya, Eric Ni, Eric Noland, Geng Yan, George Tucker, George{-}Cristian Muraru, Grigory Rozhdestvenskiy, Henryk Michalewski, Ian Tenney, Ivan Grishchenko, Jacob Austin, James Keeling, Jane Labanowski, Jean{-}Baptiste Lespiau, Jeff Stanway, Jenny Brennan, Jeremy Chen, Johan Ferret, Justin Chiu, and et~al.
\newblock Gemma: Open models based on gemini research and technology.
\newblock \emph{CoRR}, abs/2403.08295, 2024.
\newblock \doi{10.48550/ARXIV.2403.08295}.
\newblock URL \url{https://doi.org/10.48550/arXiv.2403.08295}.

\bibitem[Mostafazadeh et~al.(2016)Mostafazadeh, Chambers, He, Parikh, Batra, Vanderwende, Kohli, and Allen]{storycloze}
Nasrin Mostafazadeh, Nathanael Chambers, Xiaodong He, Devi Parikh, Dhruv Batra, Lucy Vanderwende, Pushmeet Kohli, and James Allen.
\newblock A corpus and cloze evaluation for deeper understanding of commonsense stories.
\newblock In Kevin Knight, Ani Nenkova, and Owen Rambow (eds.), \emph{Proceedings of the 2016 Conference of the North {A}merican Chapter of the Association for Computational Linguistics: Human Language Technologies}, pp.\  839--849, San Diego, California, June 2016. Association for Computational Linguistics.
\newblock \doi{10.18653/v1/N16-1098}.
\newblock URL \url{https://aclanthology.org/N16-1098}.

\bibitem[Muennighoff et~al.(2023)Muennighoff, Wang, Sutawika, Roberts, Biderman, Scao, Bari, Shen, Yong, Schoelkopf, Tang, Radev, Aji, Almubarak, Albanie, Alyafeai, Webson, Raff, and Raffel]{BLOOMZ}
Niklas Muennighoff, Thomas Wang, Lintang Sutawika, Adam Roberts, Stella Biderman, Teven~Le Scao, M.~Saiful Bari, Sheng Shen, Zheng~Xin Yong, Hailey Schoelkopf, Xiangru Tang, Dragomir Radev, Alham~Fikri Aji, Khalid Almubarak, Samuel Albanie, Zaid Alyafeai, Albert Webson, Edward Raff, and Colin Raffel.
\newblock Crosslingual generalization through multitask finetuning.
\newblock In Anna Rogers, Jordan~L. Boyd{-}Graber, and Naoaki Okazaki (eds.), \emph{Proceedings of the 61st Annual Meeting of the Association for Computational Linguistics (Volume 1: Long Papers), {ACL} 2023, Toronto, Canada, July 9-14, 2023}, pp.\  15991--16111. Association for Computational Linguistics, 2023.
\newblock \doi{10.18653/V1/2023.ACL-LONG.891}.
\newblock URL \url{https://doi.org/10.18653/v1/2023.acl-long.891}.

\bibitem[OpenAI(2023)]{gpt4}
OpenAI.
\newblock {GPT-4} technical report.
\newblock \emph{CoRR}, abs/2303.08774, 2023.
\newblock \doi{10.48550/ARXIV.2303.08774}.
\newblock URL \url{https://doi.org/10.48550/arXiv.2303.08774}.

\bibitem[Park et~al.(2024)Park, Choe, and Veitch]{linear}
Kiho Park, Yo~Joong Choe, and Victor Veitch.
\newblock The linear representation hypothesis and the geometry of large language models.
\newblock In \emph{Forty-first International Conference on Machine Learning, {ICML} 2024, Vienna, Austria, July 21-27, 2024}. OpenReview.net, 2024.
\newblock URL \url{https://openreview.net/forum?id=UGpGkLzwpP}.

\bibitem[Ponti et~al.(2020)Ponti, Glavas, Majewska, Liu, Vulic, and Korhonen]{xcopa}
Edoardo~Maria Ponti, Goran Glavas, Olga Majewska, Qianchu Liu, Ivan Vulic, and Anna Korhonen.
\newblock {XCOPA:} {A} multilingual dataset for causal commonsense reasoning.
\newblock In Bonnie Webber, Trevor Cohn, Yulan He, and Yang Liu (eds.), \emph{Proceedings of the 2020 Conference on Empirical Methods in Natural Language Processing, {EMNLP} 2020, Online, November 16-20, 2020}, pp.\  2362--2376. Association for Computational Linguistics, 2020.
\newblock \doi{10.18653/V1/2020.EMNLP-MAIN.185}.
\newblock URL \url{https://doi.org/10.18653/v1/2020.emnlp-main.185}.

\bibitem[Rimsky et~al.(2023)Rimsky, Gabrieli, Schulz, Tong, Hubinger, and Turner]{steer-vector}
Nina Rimsky, Nick Gabrieli, Julian Schulz, Meg Tong, Evan Hubinger, and Alexander~Matt Turner.
\newblock Steering llama 2 via contrastive activation addition.
\newblock \emph{CoRR}, abs/2312.06681, 2023.
\newblock \doi{10.48550/ARXIV.2312.06681}.
\newblock URL \url{https://doi.org/10.48550/arXiv.2312.06681}.

\bibitem[Sakaguchi et~al.(2020)Sakaguchi, Bras, Bhagavatula, and Choi]{winogrande}
Keisuke Sakaguchi, Ronan~Le Bras, Chandra Bhagavatula, and Yejin Choi.
\newblock Winogrande: An adversarial winograd schema challenge at scale.
\newblock In \emph{The Thirty-Fourth {AAAI} Conference on Artificial Intelligence, {AAAI} 2020, The Thirty-Second Innovative Applications of Artificial Intelligence Conference, {IAAI} 2020, The Tenth {AAAI} Symposium on Educational Advances in Artificial Intelligence, {EAAI} 2020, New York, NY, USA, February 7-12, 2020}, pp.\  8732--8740. {AAAI} Press, 2020.
\newblock \doi{10.1609/AAAI.V34I05.6399}.
\newblock URL \url{https://doi.org/10.1609/aaai.v34i05.6399}.

\bibitem[Shin et~al.(2020)Shin, Razeghi, IV, Wallace, and Singh]{prompt-engineering}
Taylor Shin, Yasaman Razeghi, Robert L.~Logan IV, Eric Wallace, and Sameer Singh.
\newblock Autoprompt: Eliciting knowledge from language models with automatically generated prompts.
\newblock In Bonnie Webber, Trevor Cohn, Yulan He, and Yang Liu (eds.), \emph{Proceedings of the 2020 Conference on Empirical Methods in Natural Language Processing, {EMNLP} 2020, Online, November 16-20, 2020}, pp.\  4222--4235. Association for Computational Linguistics, 2020.
\newblock \doi{10.18653/V1/2020.EMNLP-MAIN.346}.
\newblock URL \url{https://doi.org/10.18653/v1/2020.emnlp-main.346}.

\bibitem[Socher et~al.(2013)Socher, Perelygin, Wu, Chuang, Manning, Ng, and Potts]{sst}
Richard Socher, Alex Perelygin, Jean Wu, Jason Chuang, Christopher~D. Manning, Andrew Ng, and Christopher Potts.
\newblock Recursive deep models for semantic compositionality over a sentiment treebank.
\newblock In \emph{Proceedings of the 2013 Conference on Empirical Methods in Natural Language Processing}, pp.\  1631--1642, Seattle, Washington, USA, October 2013. Association for Computational Linguistics.
\newblock URL \url{https://www.aclweb.org/anthology/D13-1170}.

\bibitem[Subramani et~al.(2022)Subramani, Suresh, and Peters]{engineering1}
Nishant Subramani, Nivedita Suresh, and Matthew~E. Peters.
\newblock Extracting latent steering vectors from pretrained language models.
\newblock In Smaranda Muresan, Preslav Nakov, and Aline Villavicencio (eds.), \emph{Findings of the Association for Computational Linguistics: {ACL} 2022, Dublin, Ireland, May 22-27, 2022}, pp.\  566--581. Association for Computational Linguistics, 2022.
\newblock \doi{10.18653/V1/2022.FINDINGS-ACL.48}.
\newblock URL \url{https://doi.org/10.18653/v1/2022.findings-acl.48}.

\bibitem[Touvron et~al.(2023)Touvron, Martin, Stone, Albert, Almahairi, Babaei, Bashlykov, Batra, Bhargava, Bhosale, Bikel, Blecher, Canton{-}Ferrer, Chen, Cucurull, Esiobu, Fernandes, Fu, Fu, Fuller, Gao, Goswami, Goyal, Hartshorn, Hosseini, Hou, Inan, Kardas, Kerkez, Khabsa, Kloumann, Korenev, Koura, Lachaux, Lavril, Lee, Liskovich, Lu, Mao, Martinet, Mihaylov, Mishra, Molybog, Nie, Poulton, Reizenstein, Rungta, Saladi, Schelten, Silva, Smith, Subramanian, Tan, Tang, Taylor, Williams, Kuan, Xu, Yan, Zarov, Zhang, Fan, Kambadur, Narang, Rodriguez, Stojnic, Edunov, and Scialom]{llama}
Hugo Touvron, Louis Martin, Kevin Stone, Peter Albert, Amjad Almahairi, Yasmine Babaei, Nikolay Bashlykov, Soumya Batra, Prajjwal Bhargava, Shruti Bhosale, Dan Bikel, Lukas Blecher, Cristian Canton{-}Ferrer, Moya Chen, Guillem Cucurull, David Esiobu, Jude Fernandes, Jeremy Fu, Wenyin Fu, Brian Fuller, Cynthia Gao, Vedanuj Goswami, Naman Goyal, Anthony Hartshorn, Saghar Hosseini, Rui Hou, Hakan Inan, Marcin Kardas, Viktor Kerkez, Madian Khabsa, Isabel Kloumann, Artem Korenev, Punit~Singh Koura, Marie{-}Anne Lachaux, Thibaut Lavril, Jenya Lee, Diana Liskovich, Yinghai Lu, Yuning Mao, Xavier Martinet, Todor Mihaylov, Pushkar Mishra, Igor Molybog, Yixin Nie, Andrew Poulton, Jeremy Reizenstein, Rashi Rungta, Kalyan Saladi, Alan Schelten, Ruan Silva, Eric~Michael Smith, Ranjan Subramanian, Xiaoqing~Ellen Tan, Binh Tang, Ross Taylor, Adina Williams, Jian~Xiang Kuan, Puxin Xu, Zheng Yan, Iliyan Zarov, Yuchen Zhang, Angela Fan, Melanie Kambadur, Sharan Narang, Aur{\'{e}}lien Rodriguez, Robert Stojnic, Sergey Edunov,
  and Thomas Scialom.
\newblock Llama 2: Open foundation and fine-tuned chat models.
\newblock \emph{CoRR}, abs/2307.09288, 2023.
\newblock \doi{10.48550/ARXIV.2307.09288}.
\newblock URL \url{https://doi.org/10.48550/arXiv.2307.09288}.

\bibitem[Turner et~al.(2023)Turner, Thiergart, Udell, Leech, Mini, and MacDiarmid]{addition}
Alexander~Matt Turner, Lisa Thiergart, David Udell, Gavin Leech, Ulisse Mini, and Monte MacDiarmid.
\newblock Activation addition: Steering language models without optimization.
\newblock \emph{CoRR}, abs/2308.10248, 2023.
\newblock \doi{10.48550/ARXIV.2308.10248}.
\newblock URL \url{https://doi.org/10.48550/arXiv.2308.10248}.

\bibitem[Wang et~al.(2023)Wang, Haddow, and Birch]{wang2023retrieval}
Weixuan Wang, Barry Haddow, and Alexandra Birch.
\newblock Retrieval-augmented multilingual knowledge editing.
\newblock \emph{arXiv preprint arXiv:2312.13040}, 2023.

\bibitem[Wang et~al.(2024{\natexlab{a}})Wang, Haddow, Birch, and Peng]{wang2024assessing}
Weixuan Wang, Barry Haddow, Alexandra Birch, and Wei Peng.
\newblock Assessing factual reliability of large language model knowledge.
\newblock In \emph{Proceedings of the 2024 Conference of the North American Chapter of the Association for Computational Linguistics: Human Language Technologies (Volume 1: Long Papers)}, pp.\  805--819, 2024{\natexlab{a}}.

\bibitem[Wang et~al.(2024{\natexlab{b}})Wang, Haddow, Peng, and Birch]{sharing}
Weixuan Wang, Barry Haddow, Wei Peng, and Alexandra Birch.
\newblock Sharing matters: Analysing neurons across languages and tasks in llms.
\newblock \emph{CoRR}, abs/2406.09265, 2024{\natexlab{b}}.
\newblock \doi{10.48550/ARXIV.2406.09265}.
\newblock URL \url{https://doi.org/10.48550/arXiv.2406.09265}.

\bibitem[Wang et~al.(2024{\natexlab{c}})Wang, Wu, Haddow, and Birch]{wang2024bridging}
Weixuan Wang, Minghao Wu, Barry Haddow, and Alexandra Birch.
\newblock Bridging the language gaps in large language models with inference-time cross-lingual intervention.
\newblock \emph{arXiv preprint arXiv:2410.12462}, 2024{\natexlab{c}}.

\bibitem[Wei et~al.(2022)Wei, Bosma, Zhao, Guu, Yu, Lester, Du, Dai, and Le]{sft}
Jason Wei, Maarten Bosma, Vincent~Y. Zhao, Kelvin Guu, Adams~Wei Yu, Brian Lester, Nan Du, Andrew~M. Dai, and Quoc~V. Le.
\newblock Finetuned language models are zero-shot learners.
\newblock In \emph{The Tenth International Conference on Learning Representations, {ICLR} 2022, Virtual Event, April 25-29, 2022}. OpenReview.net, 2022.
\newblock URL \url{https://openreview.net/forum?id=gEZrGCozdqR}.

\bibitem[Zhao et~al.(2024)Zhao, Zhang, Chen, Kawaguchi, and Bing]{handle}
Yiran Zhao, Wenxuan Zhang, Guizhen Chen, Kenji Kawaguchi, and Lidong Bing.
\newblock How do large language models handle multilingualism?
\newblock \emph{CoRR}, abs/2402.18815, 2024.
\newblock \doi{10.48550/ARXIV.2402.18815}.
\newblock URL \url{https://doi.org/10.48550/arXiv.2402.18815}.

\bibitem[Zou et~al.(2023)Zou, Phan, Chen, Campbell, Guo, Ren, Pan, Yin, Mazeika, Dombrowski, Goel, Li, Byun, Wang, Mallen, Basart, Koyejo, Song, Fredrikson, Kolter, and Hendrycks]{engineering3}
Andy Zou, Long Phan, Sarah Chen, James Campbell, Phillip Guo, Richard Ren, Alexander Pan, Xuwang Yin, Mantas Mazeika, Ann{-}Kathrin Dombrowski, Shashwat Goel, Nathaniel Li, Michael~J. Byun, Zifan Wang, Alex Mallen, Steven Basart, Sanmi Koyejo, Dawn Song, Matt Fredrikson, J.~Zico Kolter, and Dan Hendrycks.
\newblock Representation engineering: {A} top-down approach to {AI} transparency.
\newblock \emph{CoRR}, abs/2310.01405, 2023.
\newblock \doi{10.48550/ARXIV.2310.01405}.
\newblock URL \url{https://doi.org/10.48550/arXiv.2310.01405}.

\end{thebibliography}
\bibliographystyle{iclr2025_conference}

\newpage
\appendix
\section{Appendix}
\label{sec:appendix}

\subsection{Description of Tasks}
\label{appendix-data}

\begin{itemize}
    \item \textbf{\copa} Each question consists of a premise and two alternatives, with the task being to choose the alternative that more plausibly has a causal relationship with the premise. 
    \item \textbf{\story} Each question requires a model to choose the correct ending to a four-sentence story for evaluating story understanding and script learning.
    \item \textbf{\nli} It is a collection of sentence pairs manually labeled for balanced classification with the labels entailment, contradiction, and neutral.
    \item \textbf{\mmlu} It is a benchmark designed to measure knowledge acquired during pretraining, covering 57 subjects across STEM, the humanities, the social sciences, and more.
    \item \textbf{\ssttwo} and \textbf{\sstfive} They are datasets used for sentiment analysis with 2 labels (negative, positive) and 5 labels (negative, somewhat negative, neutral, somewhat positive, or positive), respectively.
    \item \textbf{\boolq} It is a question answering dataset for yes/no questions where questions are naturally occurring.
    \item \textbf{\wino} It a fill-in-a-blank task with binary options, with the goal of choosing the right option for a given sentence which requires commonsense reasoning.
    \item \textbf{\trivia} It is a realistic text-based question answering dataset which includes question-answer pairs from documents collected from Wikipedia and the web. 
    \item \textbf{\truthful} It is a benchmark to measure whether a language model is truthful in generating answers to questions. The benchmark comprises 817 questions that span 38 categories, including health, law, finance and politics. \truthful includes both multiple-choice and generation tracks. The performance of multiple-choice track is gauged using multiple-choice accuracy (MC). This metric is based on the conditional probabilities of candidate answers given the question, with a successful result counted when the truthful answer ranks first. In the generation track, we use two pre-trained judge models to evaluate the truthfulness and informativeness.
    \item \textbf{\toxi} It is a dataset that contains implicitly toxic and benign sentences mentioning 13 minority groups.
    \item \textbf{\texttt{XCOPA}} A multilingual dataset, translated from the English \copa, is used to evaluate the capacity of models to transfer commonsense reasoning across languages.
\end{itemize}

\subsection{Dataset Sizes}
\label{sec:appendix-data_count}

The size of datasets for each task are described in the \autoref{tab:datasize}.
\begin{table}[h] \scriptsize
\centering
\setlength{\tabcolsep}{4pt}
\caption{\label{tab:datasize} The number of data used for identifying key elements and testing for 11 tasks.}
\begin{tabular}{cccccccccccc}
\toprule
Task      & \copa & \story & \nli  & \mmlu  & \ssttwo & \sstfive & \boolq & \wino & \trivia & \toxi & \truthful \\ \hline
\# identify & 500  & 360        & 2000 & 12178 & 1000 & 1000 & 1000  & 1000       & 1000     & 1000    & 817        \\
\# testset   & 500  & 1511       & 5000 & 12178 & 1821 & 2210 & 3270  & 1267       & 1000     & 1400    & 817       \\
\bottomrule
\end{tabular}
\end{table}

\subsection{Elements Overlap}
\label{sec:overlap}

\begin{figure}[h]
    \centering
    \includegraphics[scale=0.5]{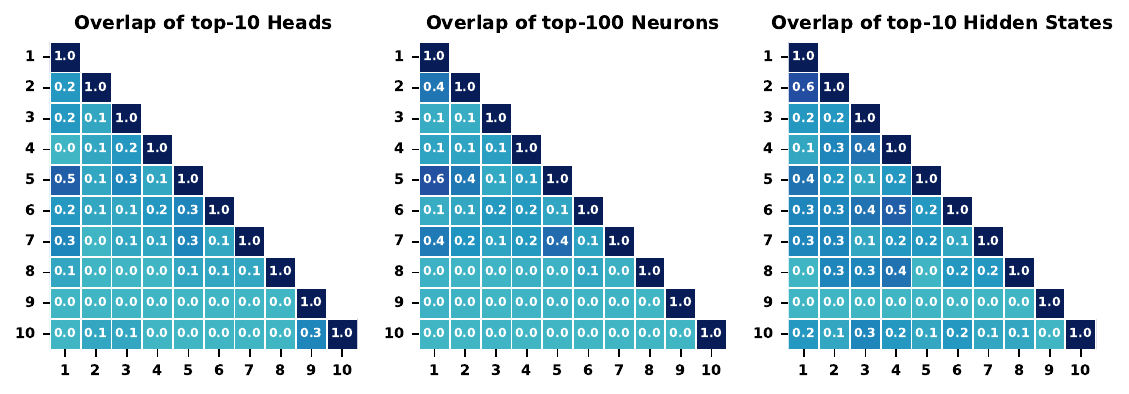}
    \caption{Overlap of identified key elements across various tasks. From 1 to 10 represents the tasks: \copa, \story, \ssttwo, \boolq, \mmlu, \nli, \wino, \trivia, \toxi, \truthful.}
    \label{task-overlap}
\end{figure}

In our analysis, detailed in \autoref{task-overlap}, we examine the overlap of identified key elements (attention heads, neurons, and hidden states) across various tasks. We observed minimal overlap across components between open-ended generation and multiple-choice tasks, particularly in attention heads and neurons. This indicates a high degree of functional specialization within these components. Unlike attention heads and neurons, hidden states demonstrate greater overlap, mainly due to their positioning in the final layer which more directly influences the model's output (as shown in \autoref{amount-contrsative}). The specialization observed among attention heads supports the notion that multi-head attention mechanisms evolve uniquely according to the task similarities. It is consistent with the finding in \citet{function-head}. 

\subsection{Cross-Lingual \sadi}
\label{appendix-crosslingual}
\begin{table*}[h] \scriptsize
\centering
\caption{\label{loacte-en-test-other} Cross-lingual results based on the identified heads/neurons from the English contrastive pairs.}
\begin{tabular}{lcccccccc}
\toprule
Language    & \textbf{id} & \textbf{it} & \textbf{sw} & \textbf{ta} & \textbf{th} & \textbf{tr} & \textbf{vi} & \textbf{zh} \\ \hline
\baseline & 51.40& 61.20& 50.20& 49.40& 50.80& 49.40& 51.80& 62.80 \\
\sft & 68.60 & 79.20 & 52.00 & 47.00 & 49.60 & 57.80 & 68.60 & 77.80 \\
\sadihead    & 62.80& 67.60& 50.40& 50.20& 51.00& 51.20& 56.60& 70.00 \\
\sadineuron  & 60.40 & 61.80 & 50.20 & 49.80 & 51.00& 59.60& 58.60& 64.00  \\
\bottomrule
\end{tabular}
\end{table*}

We have shown the effectiveness of \sadi in the multilingual scenarios where contrastive pairs, constructed in the same language as the test input, are used to identify relevant components. Further, we explored its impact in a cross-lingual setting by employing English contrastive pairs to identify key elements and then applying \sadi to multilingual test inputs. The results, along with those of \sft—which involves fine-tuning in English and testing on a multilingual dataset—are presented in Table~\ref{loacte-en-test-other}. We found that \sadi displayed enhanced language transfer capabilities, particularly in Tamil, Thai, and Turkish. These successful interventions suggest that critical elements are shared across languages, supporting \sadi's utility in cross-lingual applications.

\subsection{Activation Difference Distributions across Tasks}
\label{sec:appendix-distribution}

\begin{figure}[h]
\centering
\subfigure[Difference on \copa.]{\label{cnt-head-copa}\includegraphics[scale=0.42]{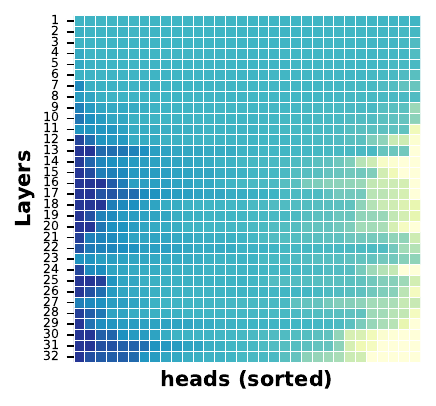}} 
\hspace{0.5em}
\subfigure[Difference on \boolq.]{\label{cnt-head-boolq}\includegraphics[scale=0.42]{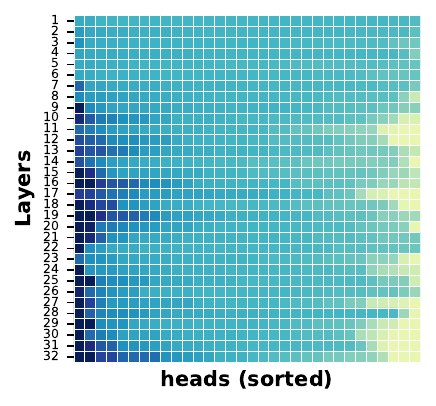}} 
\hspace{0.5em}
\subfigure[Difference on \trivia.]{\label{cnt-head-trivia}\includegraphics[scale=0.42]{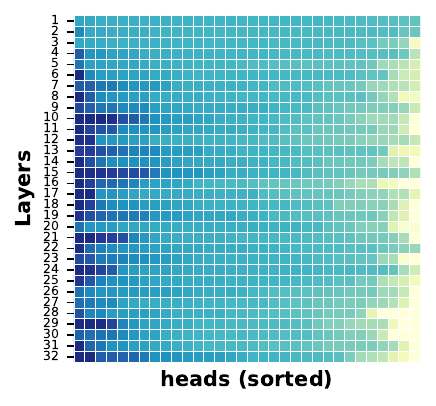}} 
\hspace{0.5em}
\subfigure[Difference on \truthful.]{\label{cnt-head-truthful}\includegraphics[scale=0.42]{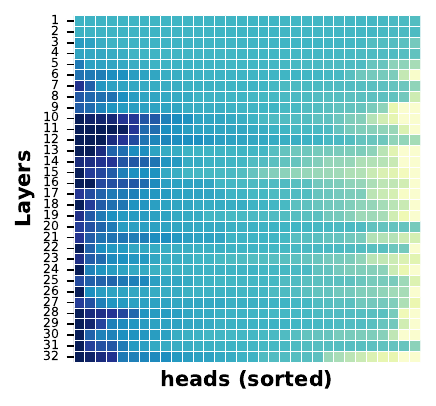}} 
\caption{\label{app-cnt-distribution-head} Activation difference of each head across layer.}
\end{figure}

\begin{figure}[h]
\centering
\subfigure[Distribution on \copa.]{\label{cnt-neuron-copa}\includegraphics[scale=0.42]{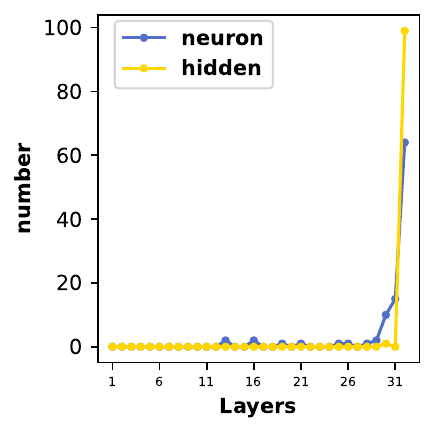}} 
\hspace{0.5em}
\subfigure[Distribution on \boolq.]{\label{cnt-neuron-boolq}\includegraphics[scale=0.42]{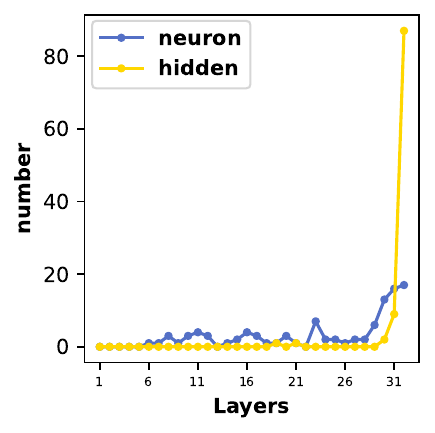}} 
\hspace{0.5em}
\subfigure[Distribution on  \trivia.]{\label{cnt-neuron-trivia}\includegraphics[scale=0.42]{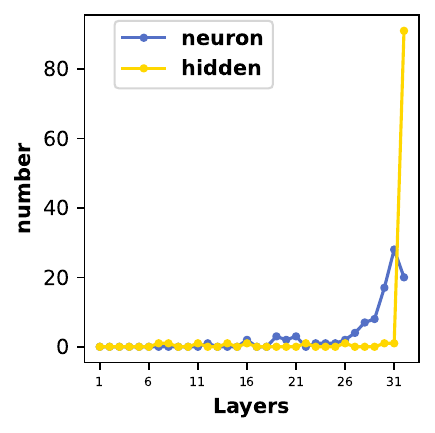}} 
\hspace{0.5em}
\subfigure[Distribution on  \truthful.]{\label{cnt-neuron-truthful}\includegraphics[scale=0.42]{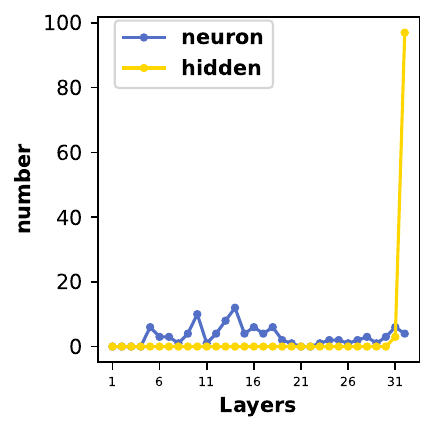}} 
\caption{\label{app-cnt-distribution-neuron} Activation difference of top-100 neurons and hidden states.}
\end{figure}

We demonstrate the distributions of activation difference across layers for heads, neurons and hidden states in \autoref{app-cnt-distribution-head} and in \autoref{app-cnt-distribution-neuron} in \copa, \boolq, \trivia, and \truthful tasks. They show consistent patterns across various tasks.

\end{document}